\documentclass[
twocolumn,
]{ceurart}

\usepackage{listings}
\usepackage[dvipsnames]{xcolor}
\usepackage{algorithm}
\usepackage{algorithmic}
\usepackage{diagbox}
\usepackage{multirow}
\usepackage{gensymb}
\usepackage[linesnumbered,ruled,vlined,algo2e]{algorithm2e}
\usepackage{indentfirst}

\SetKwInput{KwInput}{Input}                
\SetKwInput{KwOutput}{Output}  
\usepackage{setspace}

\usepackage{array}
\newcolumntype{C}[1]{>{\centering\arraybackslash}p{#1}}

\begin{document}

\copyrightyear{2023}
\copyrightclause{Copyright for this paper by its authors.
  Use permitted under Creative Commons License Attribution 4.0
  International (CC BY 4.0).}

\conference{\href{https://sites.google.com/view/sdu-aaai24/}{SDU@AAAI-24}: Workshop on Scientific Document Understanding, co-located with AAAI 2024, Vancouver, Canada}

\title{Watermark Text Pattern Spotting in Document Images}

\author[1]{Mateusz Krubiński}[email=krubinski@ufal.mff.cuni.cz]
\fnmark[1]
\cormark[1]
\author[2]{Stefan Matcovici}[email=stefmatc@amazon.com]
\cormark[1]
\author[2]{Diana Grigore}[email=digrigor@amazon.com]
\author[2]{Daniel Voinea}[email=dvoinea@amazon.com]
\author[2]{Alin-Ionut Popa}[email=popaaln@amazon.com]
\cormark[2]

\address[1]{Charles University, Faculty of Mathematics and Physics}
\address[2]{Amazon Inc.}

\fntext[1]{Work done while at Amazon.}
\cortext[1]{These authors contributed equally.}
\cortext[2]{Corresponding author.}

\begin{abstract}
Watermark text spotting in document images can offer access to an often unexplored source of information, providing crucial evidence about a record’s scope, audience and sometimes even authenticity. Stemming from the problem of text spotting, detecting and understanding watermarks in documents inherits the same hardships - in the wild, writing can come in various fonts, sizes and forms, making generic recognition a very difficult problem. To address the lack of resources in this field and propel further research, we propose a novel benchmark (\textbf{K-Watermark}) containing $65,447$ data samples generated using $\mathbf{\mathcal{W}}$\textbf{render}, a watermark text patterns rendering procedure. A validity study using humans raters yields an authenticity score of $0.51$ against pre-generated watermarked documents.
To prove the usefulness of the dataset and rendering technique, we developed an end-to-end solution ($\mathbf{\mathcal{W}}$\textbf{extract}) for detecting the bounding box instances of watermark text, while predicting the depicted text. To deal with this specific task, we introduce a variance minimization loss and a hierarchical self-attention mechanism. To the best of our knowledge, we are the first to propose an evaluation benchmark and a complete solution for retrieving watermarks from documents surpassing baselines by $5$ AP points in detection and $4$ points in character accuracy.
\end{abstract}


\maketitle

\section{Introduction}

Information  retrieval seen through the prism of visual document understanding (VDU) has recently become a mainstream task in the industry plus a hot topic in the computer vision and the natural language processing communities. It is fueled by the constant need to create automated document processing workflows and manifested through use-cases such as named-entity recognition \cite{xu2020layoutlmv2,li2021selfdoc,appalaraju2021docformer,lee2022formnet,huang2022layoutlmv3,sandu2022consent}, document classification \cite{garg2020tanda,laskar2020contextualized}, explainability in the context of validation \cite{norkute2021towards,hickmann2021analysis} or question answering \cite{lin2021bertgcn}. In spite of the growing body of research addressing VDU-related problems, several challenges remain unsolved as we are still to attain \emph{holistic understanding of the document}, to name a few - high variability in terms of layout, complexity of the visual information distribution or the multi-lingual / semantically different character of the depicted text. Additionally, since the task requires an efficient combination of visual and textual modalities that would complement each other in case one is incomplete or missing, it is difficult from the modeling perspective. 

\begin{figure}[t]
	\centering
		\includegraphics[width=0.8\linewidth]{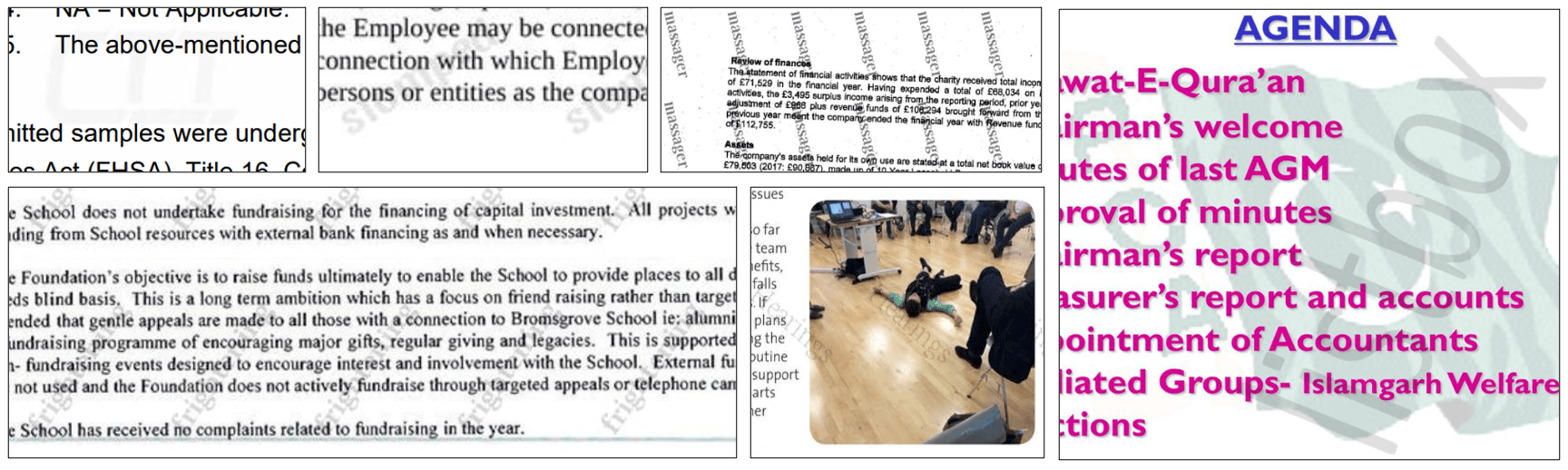}
	\caption{\textbf{Sample watermarked document image patches.} The task of watermark text recovering is challenging (a) from a visual perspective  due to resemblance with other document elements (\textit{rightmost image}) and (b) from a textual / language perspective due to high fadedness causing text misinterpretation from overlapping document text (\textit{left upper image}).}
	\label{fig:watermark_samples}
\end{figure}

\begin{figure}[t]
    \centering
    \includegraphics[width=0.99\linewidth]{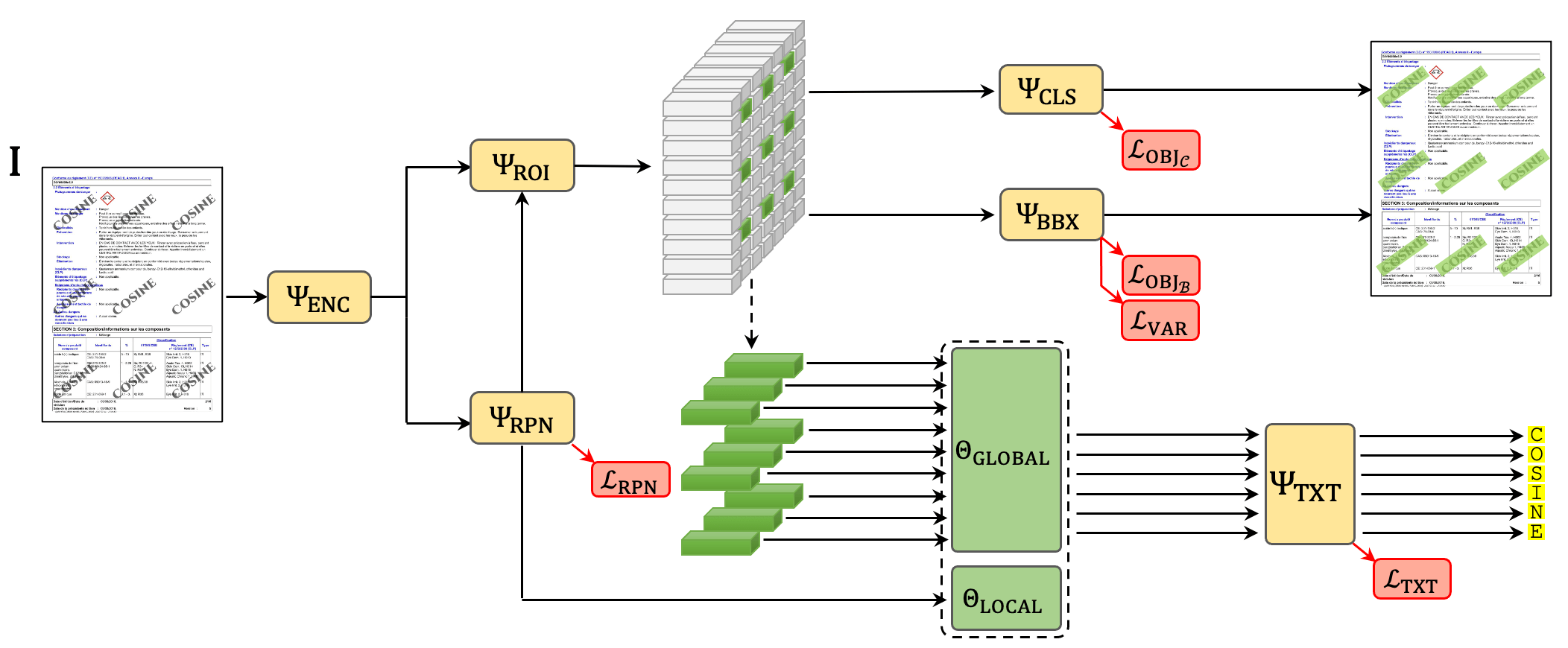}
    \caption{\textbf{Detailed overview of the proposed $\mathcal{W}$extract method.} Given an input image $\mathbf{I}$, we obtain a set of watermark region proposals via the $\mathrm{\Psi}_\texttt{CLS}$ and $\mathrm{\Psi}_\texttt{BBX}$ heads. At the same time, we construct a \emph{global} embedding representation by performing self-attention on top of the sequence of watermark region proposals generated via $\mathrm{\Psi}_\texttt{ROI}$ and a \emph{local} embedding representation built using self-attention at class-agnostic proposal feature level via $\mathrm{\Psi}_\texttt{RPN}$. Lastly, a watermark character-level prediction via $\mathrm{\Psi}_\texttt{TXT}$ is applied on top of the decoded joint global and local information.}
	\label{fig:detailed_overview}
\end{figure}

In addition to the aforementioned challenges, there are several unexplored areas of equal difficulty. One such topic is the problem of watermark text understanding from document images. Recognizing watermark content can enhance the understanding of the document information beyond the level provided by document text alone. For example, some documents might contain watermark text signaling the status of the file (\textit{e.g.}, confidential, draft) or conveying additional information regarding the discussed content  (\textit{i.e.}, hazardous substances). One document analysis domain where this is of critical importance is within compliance-related tasks where in order to demonstrate the validity of a document, one must ensure that there is no watermark text on top, and if there is one, that it follows the required pattern, \textit{e.g.}, denoting the type of document or provenience. The most difficult aspect resides in the fact that we are dealing with two pattern distributions (watermark text vs. document text) that are very similar in terms of visual and textual appearance (see Figure \ref{fig:watermark_samples}) with a high occlusion and overlapping degree between the elements. Moreover, this can be easily mistaken with other forms of text elements which can be placed diagonally inside a document, such as handwritten signatures, stamps or even text-based logos.
The main contributions of this work are: 
\begin{itemize}
    \item $\mathcal{W}$\textbf{render}, a flexible algorithmic procedure for rendering text-based patterns inside a document controlling the orientation, transparency, content and font of the watermark, enabling the creation of diverse training and evaluation datasets, 
    \item \textbf{K-Watermark}, a watermark spotting benchmark with a total of $65,447$ data samples, that is, to the best of our knowledge, the first of this kind, on which we evaluate different text spotting solutions to establish strong baselines for watermark text detection and recognition,
    \item $\mathbf{\mathcal{W}}$\textbf{extract}, an end-to-end method (see Figure \ref{fig:detailed_overview}) of detecting and recognising watermark text patterns in document images through the use of a novel loss formulation and a hierarchical self-attention encoder-decoder mechanism operating at both local and global level, achieving state-of-the-art results.
\end{itemize}

\section{Related Work}

Traditional OCR methods \cite{textract,tesseract} rarely focus on recovering text in occlusion or random orientation scenarios. Recently, a couple of lines of work \cite{ronen2022glass,liu2018fots,shi2018aster,liu2018char,zheng2020lal,baek2020character,liu2021abcnet,fang2021read,yue2020robustscanner,jaderberg2014deep} bring more attention to the task of text-spotting. Their main advantage over standard OCR techniques is their robustness with respect to background appearance variety, unknown or undefined text orientations (\textit{i.e.}, not the usual straight line orientation), unknown or mixed text fonts and even clutter with different scene elements resembling text. 

One of the most challenging aspects associated with the task of text-spotting is the ability to determine the necessary connection between the identified individual character instances such that the recovered text has the required semantic meaning. For example, the authors of \cite{baek2020character} propose a system which detects individual occurrences of characters as well as their pairwise connections (inspired by the work of \cite{cao2017realtime}) to determine the shape of the word. In our setup, when the background involves sequences of characters, the major challenge is to separate the searched text (watermark) that is occluded by or intersected with the background text.

\begin{table}[h]
    \resizebox{\columnwidth}{!}{
	\begin{tabular}{@{}c | c | c | c@{}}
		\textbf{Benchmark} & \textbf{Domain} &  \textbf{Samples} & \textbf{Supervision Level} \\
		\hline
		K-Watermark & \textit{Text Spotting} & $65,447$ & \textit{Word} \\
		DUDE \cite{van2023document} & \textit{Question Answering} & $7,947$ & \textit{Paragraph} \\
		Tobacco \cite{kumar2014structural} & \textit{Classification \& Retrieval} & $3,482$ & \textit{Document} \\
		FUNSD \cite{jaume2019} & \textit{Named-Entity Recognition} & $199$ & \textit{Word} \\
		PublayNet \cite{kumar2014structural} & \textit{Information Retrieval} & $364,232$ & \textit{Paragraph}
	\end{tabular}}
	\caption{\textbf{Common document understanding benchmarks.} Our proposed \textbf{K-Watermark} dataset falls in the category of large datasets and presents granular-level annotations (\textit{i.e.} word-level).}
	\label{tbl:benchmark_comparisons}
\end{table}

\nocite{stoian2022unstructured}
\nocite{sandu2022large}

The object detection methodology \cite{ren2015faster,redmon2016you,tan2020efficientdet,Lin_2017_ICCV,carion2020end,li2022exploring,matcovici2023k} can address the issue of localizing visual elements from text forms. The general principle is to localize and classify object regions within the image space by leveraging the local context and the discriminative aspect of the searched object classes. Our approach is based on the work of \cite{ren2015faster} combined with \cite{Lin_2017_ICCV} to help us spot watermark instances, by optimally combining modular features at different scales.
The majority of document related benchmarks (see Table \ref{tbl:benchmark_comparisons}) focus on question answering, NER or general document level reasoning. Text spotting on documents has diverging needs, the most significant being a costly and inefficient world-level supervision process, proving the need for constructing a principled data setup.

\section{K-Watermark Dataset}
\textcolor{black}{
In a real-world scenario, the watermark alteration of a document is performed using specialized document editing tools (\textit{e.g.}, Microsoft Word, Adobe Acrobat). This is a one-way prohibitive process which does not grant access to watermark text details such as size, position, word, font, angle, all of which represent necessary ground-truth information with respect to the task of watermark text spotting. Thus, one is forced to collect these details using human annotators, which turns out to be a complex and time-consuming process, as it requires a high degree of concentration to perfectly align the depicted watermark text bounding boxes and compensate for the various degrees of transparency and occlusion against visual elements. There are clear advantages of obtaining them in an automated manner as it is more efficient and frugal. Given the lack of available benchmarks on watermark text spotting from document images, we define an image-based data generation procedure called $\mathcal{W}$\textbf{render}, that augments clear document images with watermark text patterns in a fully programmatic setup allowing for full control over all the aspects of the process. Augmentation is applied in such a manner that the watermark text overlaps with the original text and other visual elements (\textit{e.g.}, tables, figures). The complexity of obtaining these examples stems from the multiple degrees of freedom involved: inserting the text, choosing the font type, the pattern density and visibility of the rotation angle all bring their own difficulty.
}

\begin{algorithm}
	\setstretch{1.1}
	\DontPrintSemicolon
	\KwInput{$\mathbf{I}_{\mathrm{DOC}} \in \mathbb{R}^{h \times w \times 3}$}
	\KwOutput{$\mathbf{I}$,  $\{b_i \; \}_{i=1}^{N_{words}}$, $\alpha$, $s$
	\newline where $\mathbf{I} \in \mathbb{R}^{h \times w \times 3}$, $b_i \in [0, 1]^4$, $s = \{s_{i}\}_{i=1}^{N_{chars}}$ with $s_{i} \in \{\texttt{a} \dots \texttt{z}\}$}
	
	$\alpha \gets$ $\texttt{Uniform}$ $(\;\frac{-\pi}{2}, \frac{\pi}{2})\;$

	$n_{words} \gets$ $\texttt{Uniform}$ $(\;1, 12\;)$ 
	
	$t$ $\gets 0.1 + 0.5 \cdot \texttt{Random}_\texttt{Beta}(\alpha=1, \; \beta = 1.5)$
	
	$f$ $\gets$ $\texttt{Uniform}$ (\texttt{GoogleFonts})
	
	$s \gets$ $\texttt{Uniform}$ (\texttt{NewsWords})
	
	$\mathbf{I} \gets \mathbf{I}_\mathrm{DOC}$

    \For{\texttt{j} $\;= 1 \dots n_{words}$}
	{
    	\For{\texttt{k} $\;= 1 \dots n_{words}$}
    	{
    	$pos = (\frac{j \cdot w}{n_{words}}, \frac{k \cdot h}{n_{words}})$
    	
    	$\mathbf{I}, b_i \gets$ \texttt{InsertWatermark}$(\mathbf{I}, s, t, f, pos$)
    	}
    }
	\caption{$\mathbf{\mathcal{W}}render$}
	\label{alg:wrender}
\end{algorithm}

\subsection{Wrender}
\textcolor{black}{
 In order to make the text insertion procedure as realistic as possible, we studied the typical watermark patterns (based on samples from the web and historical data), and concluded that they are rendered in a grid-like pattern, with the same text, color, font, angular placement and transparency. Their position does not take into consideration other document elements such as text or tables. Thus, our watermark insertion technique $\mathbf{\mathcal{W}render}$, was designed with these real-world observed constraints in mind. Algorithm \ref{alg:wrender} receives as input a document image $\mathbf{I}_\mathrm{DOC}$ and returns the watermarked image $\mathbf{I}$ together with a list $\{(b_i \;, \; \alpha_i \;,\; s_i)\}_{i=1}^{N_{words}}$ containing the normalized bounding box coordinates $b_i \in [0, 1]^4$ of the inserted watermark text, their angle $\alpha_i \in (-\frac{\pi}{2}, \frac{\pi}{2})$ with respect to the horizontal axis as well as the watermark text $s = \{s_{i}\}_{i=1}^{N_{chars}}$ with $s_{i} \in \{\texttt{a} \dots \texttt{z}\}$.
 }
 
 \begin{figure}[!h]
    \centering
    \setlength{\tabcolsep}{5pt}
    \includegraphics[scale=0.14]{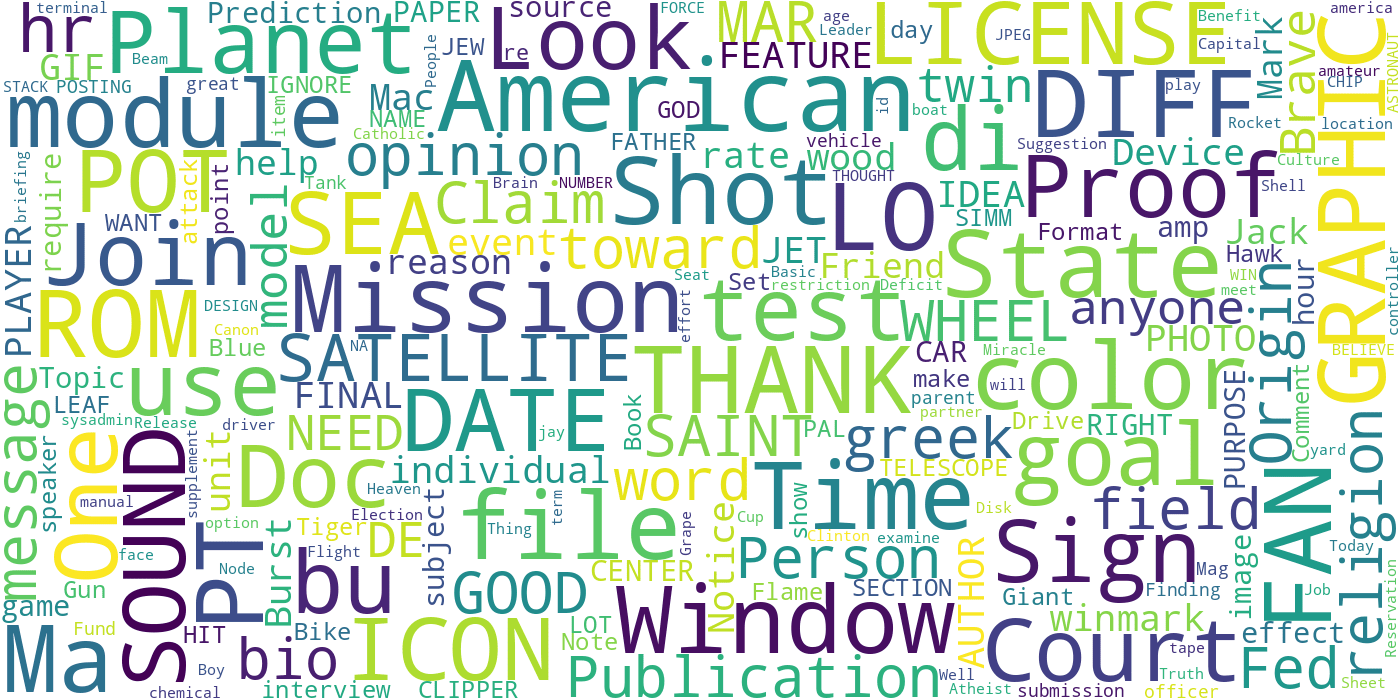}
    \caption{Wordcloud visualization encoding a display of the frequency of the words used throughout our training dataset. These are some of the most frequent english words commonly used in newspapers and mass media content.}
	\label{fig:train_wordcloud}
\end{figure}

\begin{figure*}[!h]
    \centering
    \setlength{\tabcolsep}{5pt}
    \includegraphics[width=0.83\linewidth]{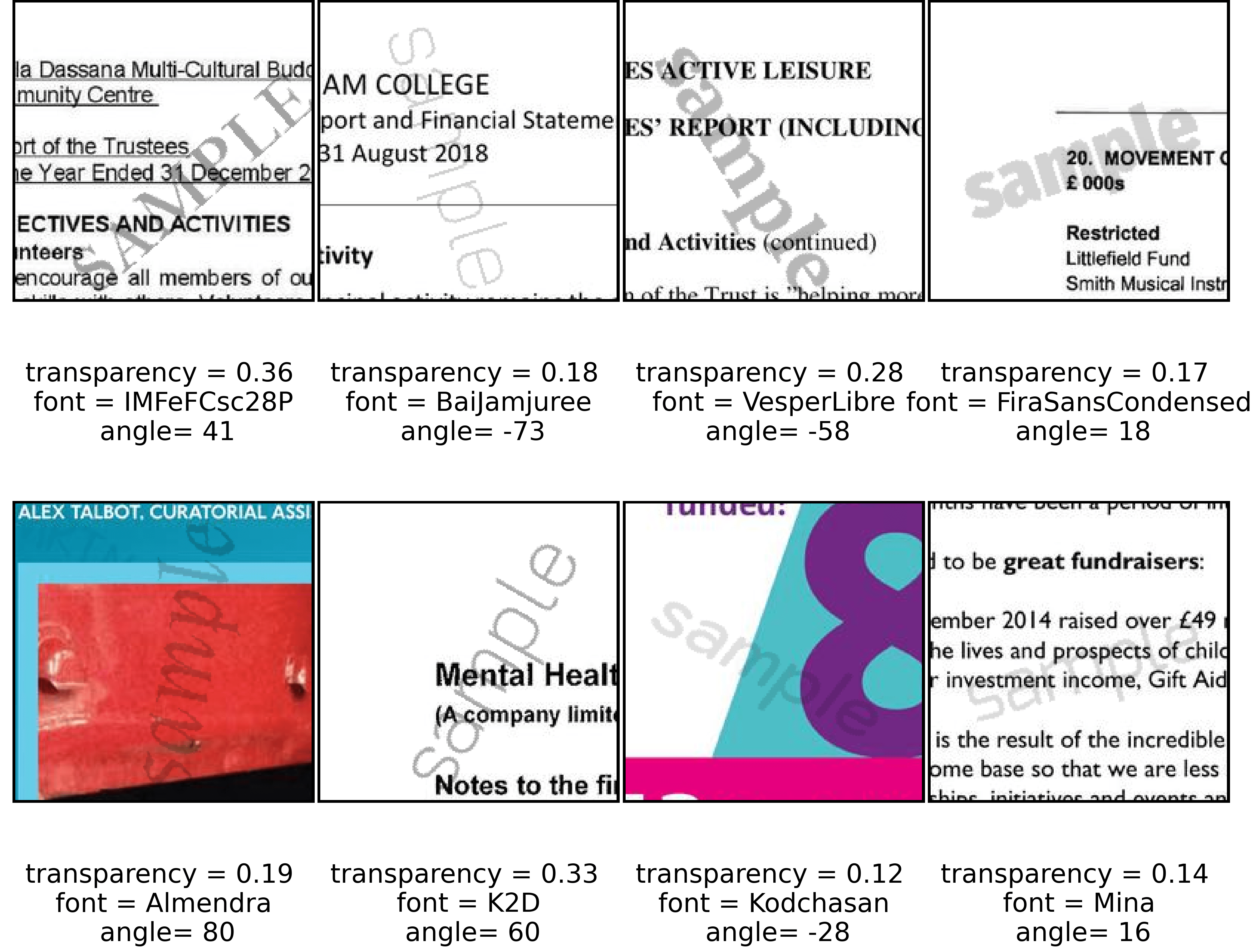}
    \caption{\textbf{Generated individual watermark text patterns.} Using our $\mathcal{W}$\textbf{render} insertion procedure we have full control over the watermark pattern insertion process similar to off-the-shelf professional document editing tools. Bellow each image patch, we specify the transparency (\textit{i.e.} on a scale from $0$ to $1$), font (\textit{i.e.} randomly sampled from \cite{google}) and angle (\textit{i.e.} from $-90 \degree$ to $90 \degree$ degrees) used. }
	\label{fig:train_watermarks}
\end{figure*}

\textcolor{black}{
For each image, the number of watermark text bounding boxes is chosen random from the list of perfect square numbers between $1$ and $144$. The reason is to have a grid positioning of the inserted words according to the square root of the selected number (\textit{e.g.}, for a value of $16$ it will create a grid $4 \times 4$ of watermark words). Also, when we position the words on the determined grid, we added a small offset to the entire grid with respect to the page size so that we do not have the exact word positioning for $2$ pages with the same grid size. Since based on our observations all the watermarks have the same angular orientation, $\alpha$ and the same text $s$, we imposed this constraint when rendering the watermarks on the grid. Moreover, the watermark text is usually very faded. Thus, we constrained the visibility of the inserted text to follow a Beta distribution which generates a higher number of values closer to the lower bound of $0.1$ and a fewer number of values closer to the upper bound of $0.6$. Basically, a value closer to $0.1$ implies a higher fadedness and as it gets closer to $0.6$, it becomes more visible. For the font type, we randomly sampled from the Google Fonts database \cite{google}.}

\textcolor{black}{
In order to apply our augmentation technique, we use clean document images, not containing watermarks. For illustration purposes, we make use of the publicly available multi-page document datasets Kleister NDA and Kleister Charity \cite{stanislawek2021kleister}, using the documents as background images to paste the watermark text patterns. As a consequence, we named the resulting dataset \textbf{K-Watermark}. The document data is split at the page level, resulting in the following split: $57,947$ images for train, $2,500$ images for validation and $5,000$ images for test. For the test and the validation data, we have drawn real words from \cite{newsgroup20} and we have used different fonts than those used for training split, as well as different text densities of the watermark text. Additionally, when generating the test / validation set, we considered the situation of no watermark words, to include document scenarios with no inserted watermark to better estimate the false positive rate. We will release the \textbf{K-Watermark} dataset together with the code that renders patterns thus encouraging the research on document image watermarks in more diverse scenarios without the explicit need for labeling large amounts of data. Furthermore, the watermark insertion procedure can be used dinamically during training, in order to reduce the possible overfitting, the augmentated data might induce.}

\textcolor{black}{
To prove the validity of our hypotheses about real-world watermarks we conducted a human perception study. We sampled $100$ images watermarked using $\mathcal{W}$\textbf{render} and $100$ random watermark documents collected using web search (\textit{i.e.}, pre-generated in the wild). Next, we asked $5$ human annotators to rank them as $\mathcal{W}$\textbf{} or pre-generated. The outcome was $0.51$ and $0.49$ F1-score for the $\mathcal{W}$\textbf{render} and pre-generated class, respectively, proving that our procedure is able to mimic the watermark insertion procedure from specialized tools successfully and enabling us to obtain text annotations (bonding box localization and depicted text) with far less resources.}

During the training procedure, \textbf{K-Watermark} samples are dynamically generated using $\mathcal{W}$\textbf{render} by retrieving random words from \cite{newsgroup20} and placing them as watermark patterns on top of the document page. A word-cloud visualization is illustrated in Figure \ref{fig:train_wordcloud} with words used as watermark text patterns during training. These are frequent words from the English vocabulary of various lengths and with different semantic meanings. In Table \ref{tbl:kwatermark_statistics} we illustrate word level statistics for the validation and test sets emphasizing the high density of watermark text patterns at both dataset and page level as well as the high overlap against document text.

In Figure \ref{fig:train_watermarks} we illustrate samples generated dynamically by using the $\mathcal{W}$\textbf{render} procedure. It is noticeable how the watermark text augmentation is impacted by the multiple degrees of freedom across the angle, font and fadedness elements. By doing this, we increase the generability potential of our proposed approach. 

\begin{table*}[t]
	\centering
	\begin{tabular}{C{20mm}|C{20mm}|C{20mm}|C{20mm}|C{20mm}|C{20mm}}
		\multirow{2}{*}{\textbf{Split Type}} & \multicolumn{3}{c|}{$\#$ \textbf{of Watermarks}} & \multicolumn{2}{c}{\textbf{Document Text Overlap}} \\
		\cline{2-6}
		 & \textbf{Total} &  \textbf{Mean} & \textbf{Median} &  \textbf{Mean} & \textbf{Median} \\
 		\hline
		Validation & $182,325$ & $72.93 \pm 62.564$ & $49.0$ & $53.56 \pm 23.76$ & $56.2$\\
		Test & $360,973$ & $72.19 \pm 62.26$ & $49.0$ & $54.4 \pm 23.39$ & $57.3$ \\
	\end{tabular}
	\caption{\textbf{Common statistics for validation and test subsets.} We emphasize the high density of watermark text pattern content at dataset and page level. Moreover, in columns $5$ and $6$ we highlight high overlap between document and watermark text which increases the complexity of the proposed task.}
	\label{tbl:kwatermark_statistics}
\end{table*}

\section{Methodology}
Given an image $\mathbf{I} \in \mathbb{R}^{h \times w \times 3}$, the objective is to predict a watermark text string $\tilde{s} = \{\tilde{s}_{i}\}_{i=1}^{N_{chars}}$ with $\tilde{s}_{i} \in \{\texttt{a} \dots \texttt{z}\}$ and a list of watermark image regions defined by the tuples $\{(\tilde{b}_i \;, \; \tilde{\alpha}_i)\}_{i=1}^{N_{p}}$, where $\tilde{b}_i \in [0, 1]^4$ represents the normalized bounding box coordinates with respect to image height and width, $h$ and $w$, respectively, $\tilde{\alpha}_i \in (-\pi/2, \pi/2)$ represents the angular orientation of the watermark text. In practice, the angles and the watermark text from all the ground truth bounding boxes are the same. As backbone for our proposed framework, $\mathcal{W}$\textbf{extract}, we use Faster-RCNN \cite{ren2015faster} combined with feature pyramid networks \cite{Lin_2017_ICCV}

\begin{table*}[t]
	\centering
	\begin{tabular}{C{31mm}|C{14mm}|C{14mm}|C{14mm}|C{14mm}|C{14mm}|C{14mm}}
		\textbf{Model} &  mAP & AP@$50$ & AP@$75$ & mAR & AR$@50$ & AR$@75$ \\
		\hline
	    TextSnake \cite{long2018textsnake} & $6$ & $20$ & $2$ & $15$ & $36$ & $10$ \\
		CRAFT \cite{baek2020character} & $7$ & $26$ & $1$ & $34$ & $46$ & $8$ \\
        DBNet++ \cite{liao2022real} & $10$ & $29$ & $4$ & $16$ & $38$ & $11$ \\
		UNITS \cite{kil2023towards} & $8$ & $15$ & $7$ & $22$ & $38$ & $23$ \\
		TCM \cite{yu2023turning} & $14$ & $42$ & $5$ & $20$ & $49$ & $12$ \\
		\hline
	    \textit{Fine-tuned} TextSnake \cite{long2018textsnake} & $32$ & $82$ & $14$ & $44$ & $91$ & $36$ \\
		\textit{Fine-tuned} DBNet++ \cite{liao2022real} & $48$ & $92$ & $44$ & $60$ & $\mathbf{97}$ & $67$ \\
		\textit{Fine-tuned} UNITS \cite{kil2023towards} & $50$ & $95$ & $48$ & $62$ & $\mathbf{97}$ & $68$ \\
		\textit{Fine-tuned} TCM \cite{yu2023turning} & $52$ & $95$ & $49$ & $58$ & $96$ & $63$ \\
		\hline 
		$\mathcal{W}$\textbf{extract} w/o $\mathcal{L}_\texttt{VAR}$ & $56$ & $\mathbf{96}$ & $58$ & $61$ & $96$ & $69$ \\
		$\mathcal{W}$\textbf{extract} & $\mathbf{58}$ & $\mathbf{96}$ & $\mathbf{65}$ & $\mathbf{63}$ & $\mathbf{97}$ & $\mathbf{72}$ \\
	\end{tabular}
	\caption{\textbf{Watermark Text Detection Results on K-Watermark Test Set.} We compare against state-of-the-art baselines for text spotting. First baseline (lines $2-6$) is by running the off-the-shelf version of these approaches on the raw documents (pre-watermark) and the watermarked documents (post-watermark) to retain the watermark text only via prediction set differentiation. The second baseline (lines $7-10$) is to fine-tune (when possible) on K-Watermark and evaluate directly.}
	\label{tbl:detection_results}
\end{table*}

\begin{align}
	\mathrm{\Psi}_{\texttt{OBJ}_\mathcal{C}} (\cdot) &= \mathrm{\Psi}_\texttt{CLS} \circ \mathrm{\Psi}_\texttt{ROI} \circ \mathrm{\Psi}_\texttt{RPN} \circ \mathrm{\Psi}_\texttt{ENC} (\cdot) \label{eq:cls} \\
	\mathrm{\Psi}_{\texttt{OBJ}_\mathcal{B}} (\cdot) &= \mathrm{\Psi}_\texttt{BBX} \circ \mathrm{\Psi}_\texttt{ROI} \circ \mathrm{\Psi}_\texttt{RPN} \circ \mathrm{\Psi}_\texttt{ENC} (\cdot) \label{eq:regr}
\end{align}

\noindent where $\mathrm{\Psi}_\texttt{ENC}$ is represented by a common image encoder which in our case is \texttt{ResNet50} \cite{He2015}, $\mathrm{\Psi}_\texttt{RPN}$ is a class-agnostic region proposal network (RPN), $\mathrm{\Psi}_\texttt{ROI}$ is a region of interest (ROI) feature encoder which links the proposals from $\mathrm{\Psi}_\texttt{RPN}$. $\mathrm{\Psi}_{\texttt{CLS}}$ and $\mathrm{\Psi}_{\texttt{BBX}}$ represent classification and regression heads, respectively, designed to align the region proposals received from the $\mathrm{\Psi}_\texttt{RPN}$ with the watermark class type and their exact corresponding bounding boxes within the image. In our case, the $\mathrm{\Psi}_{\texttt{CLS}}$ is a binary classifier stating whether the proposed bounding boxes depict watermark text or not. Moreover, the regression head $\mathrm{\Psi}_{\texttt{BBX}}$ not only predicts the bounding boxes of the watermark regions, but also the orientation of the detected watermark text patterns.

\begin{table}[t]
	\centering
	\begin{tabular}{C{28mm}|C{14mm}|C{14mm}}
		\textbf{Backbone} &  mAP & mAR \\
		\hline
		\texttt{ResNet}$50$ \cite{He2015} & $\mathbf{58}$ & $\mathbf{63}$  \\
		\texttt{SWIN-Base} \cite{liu2021swin} & $51$ & $53$  \\
		\texttt{ViTDET-Base} \cite{li2022exploring} & $53$ & $55$  \\
		\hline
	\end{tabular}
	\caption{\textbf{Detection Results with Various Image Encoding Backbones.} We experimented with different image encoding backbones to validate the best encoding architecture to proceed further with the end-to-end watermark text spotting task.}
	\label{tbl:ablation_backbone}
\end{table}

\subsection{Variance Minimization Loss}

Traditionally, the Faster-RCNN framework requires losses for the $\mathrm{\Psi}_\texttt{RPN}$ module and the final $\mathrm{\Psi}_{\texttt{CLS}}$ / $\mathrm{\Psi}_{\texttt{BBX}}$ heads. We propose an additional loss term $\mathcal{L}_\texttt{VAR}$, which minimizes angle, width and height variations across predictions from each single-page document image. This loss is predicated on the observation that across a document, all the watermark text bounding boxes are expected to have the same width, height, and angular orientation. During training, if the cardinality of the watermark sequence is $1$, the variance loss term evaluates to $0$ (as the variance of a set with a single element is $0$), thus not affecting the training process.

\begin{align*}
	\mathcal{L}_\texttt{VAR} &=   \sum_{i=1}^{N_p} [(\tilde{b}_{i2} - \tilde{b}_{i0}) - \frac{1}{N_p} \sum_{j=1}^{N_p} (\tilde{b}_{j2} - \tilde{b}_{j0}) ]^2 \\
	&\qquad + \sum_{i=1}^{N_p} [(\tilde{b}_{i3} - \tilde{b}_{i1}) - \frac{1}{N_p} \sum_{j=1}^{N_p} (\tilde{b}_{j3} - \tilde{b}_{j1}) ]^2 \\
	&\qquad + \sum_{i=1}^{N_p} (\tilde{\alpha}_i - \frac{1}{N_p} \sum_{j=1}^{N_p} \tilde{\alpha}_j )^2 
\end{align*}

\noindent where the $3$ terms of the sum correspond respectively to the height, the width, and the angle of the predicted list of watermark regions. Additionally, $\mathcal{L}_\texttt{VAR}$ is divided by $N_p$, as a normalization factor with respect to the total number of watermark proposals. As a consequence, the total loss formulation becomes

\begin{align}
	\mathcal{L}_{\texttt{VAR}} &= \mathcal{L}_{\texttt{OBJ}_\mathcal{C}} + \mathcal{L}_{\texttt{OBJ}_\mathcal{B}} + \mathcal{L}_{\texttt{RPN}} + \mathcal{L}_{\texttt{VAR}} + \mathcal{L}_{\texttt{TXT}}
	\label{lbl:loss}
\end{align}

\noindent where $\mathcal{L}_{\texttt{VAR}}$ represents the variance minimization loss term. The $\mathcal{L}_{\texttt{TXT}}$ term represents the text recognition loss which we will define in the following section.

\subsection{Local global self-attention} \label{sec:recognition}

For the text recognition part, we propose the following hierarchical self-attention mechanism. \emph{Firstly}, we compute local feature representations based on descriptors from the $\mathrm{\Psi}_\texttt{RPN}$ head.

\begin{align*}
	\mathrm{\Theta}_\texttt{LOCAL}^i &= softmax\left(\frac{\phi^Q(\mathrm{\Psi}_\texttt{RPN}^i ) \phi^K(\mathrm{\Psi}_\texttt{RPN}^i) ^\top}{\sqrt{d_\texttt{LOCAL}}}\right) \phi^V(\mathrm{\Psi}_\texttt{RPN}^i)
\end{align*}

\noindent where $\mathrm{\Psi}_\texttt{RPN}^i$ represents the class-agnostic descriptor corresponding to the detection $i \in \{1 \dots N_{p}\}$ and $\phi(\cdot): \mathbb{R}^{d_h \times d_w \times d_\texttt{RPN}} \rightarrow \mathbb{R}^{d_h \times d_\texttt{LOCAL}}$ is a projection function based on a $1 \times 1$ convolution, followed by a reshape operation that flattens the width dimension.  The intuition is to iterate over each prediction and perform self-attention across a sequence built along the height of the descriptor tensor in order to retrieve meaningful local information with respect to the text recognition task. This is intended to mimic the heuristic of looking across the characters of a word to determine the word itself. In our experimental setup, we chose the following values for the previously listed dimensionalities: $d_h = 7$, $d_w = 7$, $d_\texttt{RPN} = 256$ and $d_\texttt{LOCAL} = 224$. The upper scripts $Q, K, V$ used for $\phi$ stand for the query, key and value projections and are associated with the attention mechanism. Finally, we obtain a sequence of local embeddings, corresponding to all the retrieved watermark text detections, $\mathrm{\Theta}_\texttt{LOCAL} = [\mathrm{\Theta}_\texttt{LOCAL}^i]_{i=1}^{N_{p}}$.

\emph{Secondly}, we compute global feature representations based on descriptors from the $\mathrm{\Psi}_\texttt{ROI}$ head. 

 \begin{align*}
 	\mathrm{\Theta}_\texttt{GLOBAL} &= softmax\left(\frac{\mathrm{\Psi}^Q_\texttt{ROI}  {\mathrm{\Psi}^K_\texttt{ROI}}^\top}{\sqrt{d_\texttt{GLOBAL}}}\right) \mathrm{\Psi}^V_\texttt{ROI}
 \end{align*}

\noindent where $\mathrm{\Theta}_\texttt{GLOBAL} \in \mathbb{R}^{N_{p} \times d_\texttt{GLOBAL}}$. The upper scripts $Q, K, V$ denote the query, key, value projections used within the attention mechanism. The intuition is to perform an implicit statistic by analysing an sequence with all the watermark detections and gather complementary information with respect to the depicted watermark text through the multi-head attention. Thus, we overcome the issues created by occlusion and fadedness of the text by looking simultaneously inside the image. 
Next, we combine both global and local feature representations into a single tensor $\mathrm{\Theta} = [\mathrm{\Theta}_\texttt{LOCAL}; \mathrm{\Theta}_\texttt{GLOBAL}]$ and project them to $\mathbb{R}^{N_{p} \times d}$, where $d$ is the embedding size of the sequence elements. By doing this, we enforce the text decoding mechanism to incorporate complementary sources of information originating from bounding box level embeddings (local) and page level embeddings (global).

\begin{figure} [!htbp]
	\begin{center}
		\scalebox{0.96}{
			\setlength{\tabcolsep}{5pt}
			\begin{tabular}{ccc}
				\includegraphics[width=0.3\linewidth]{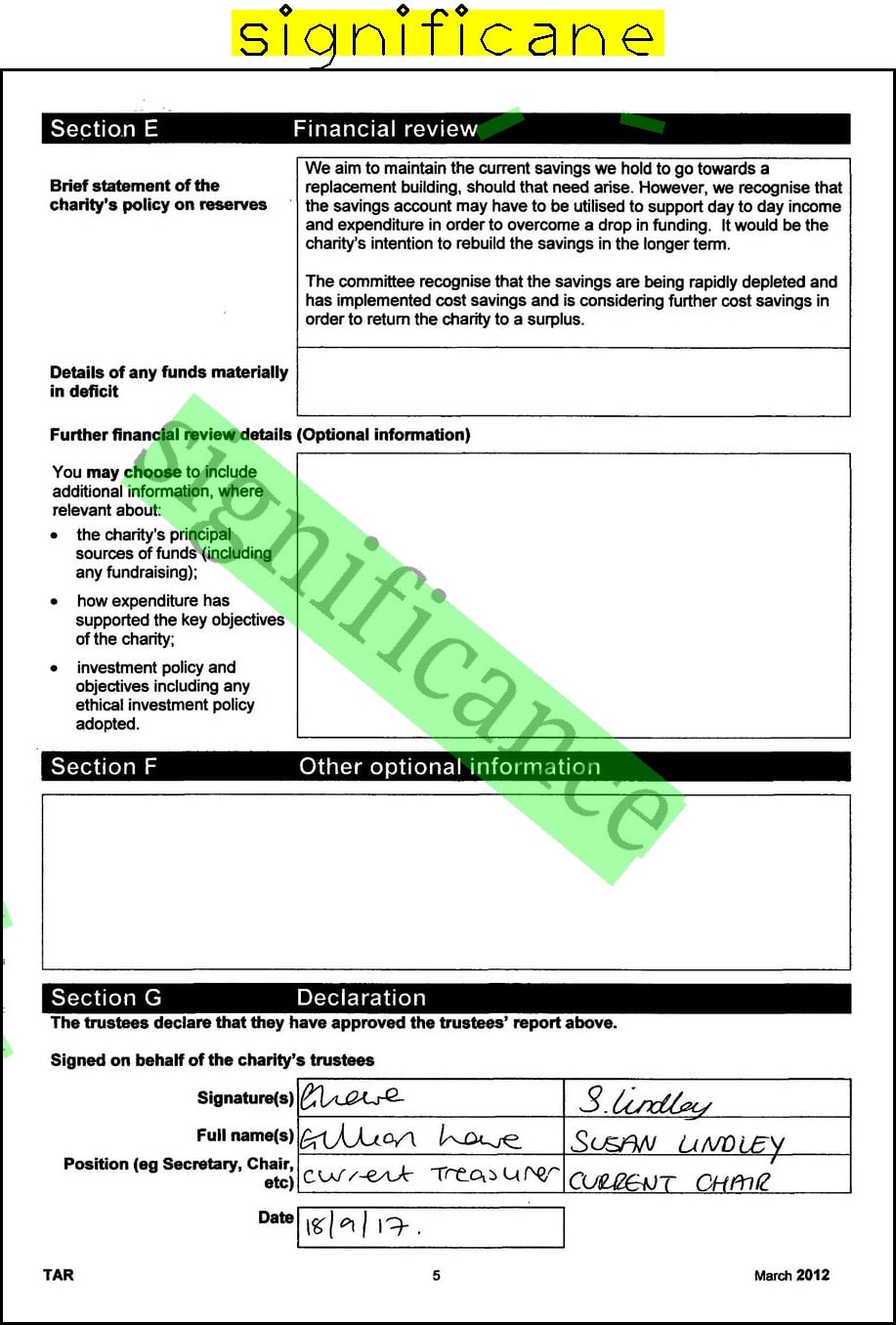} &
				\includegraphics[width=0.3\linewidth]{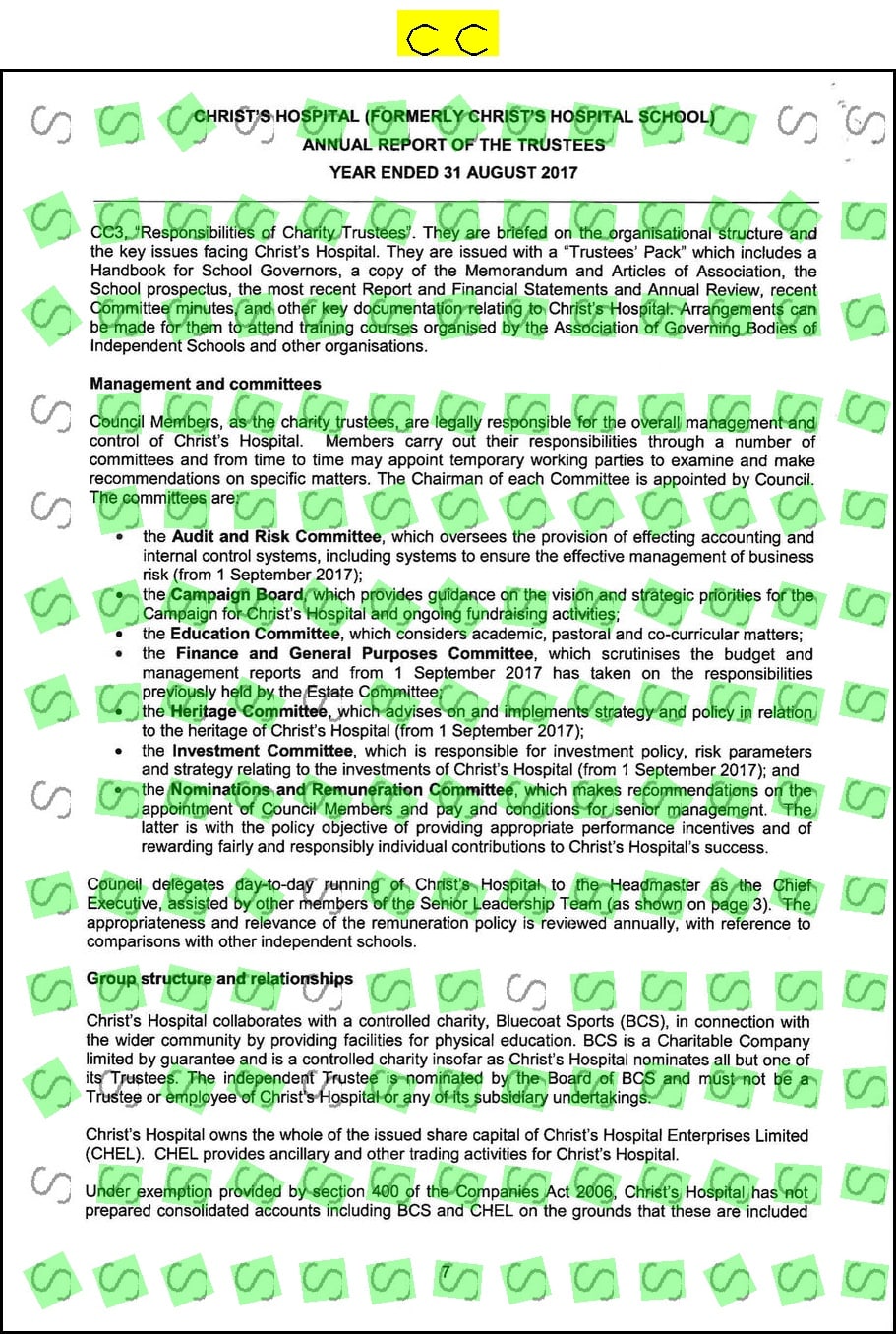} &
				\includegraphics[width=0.3\linewidth]{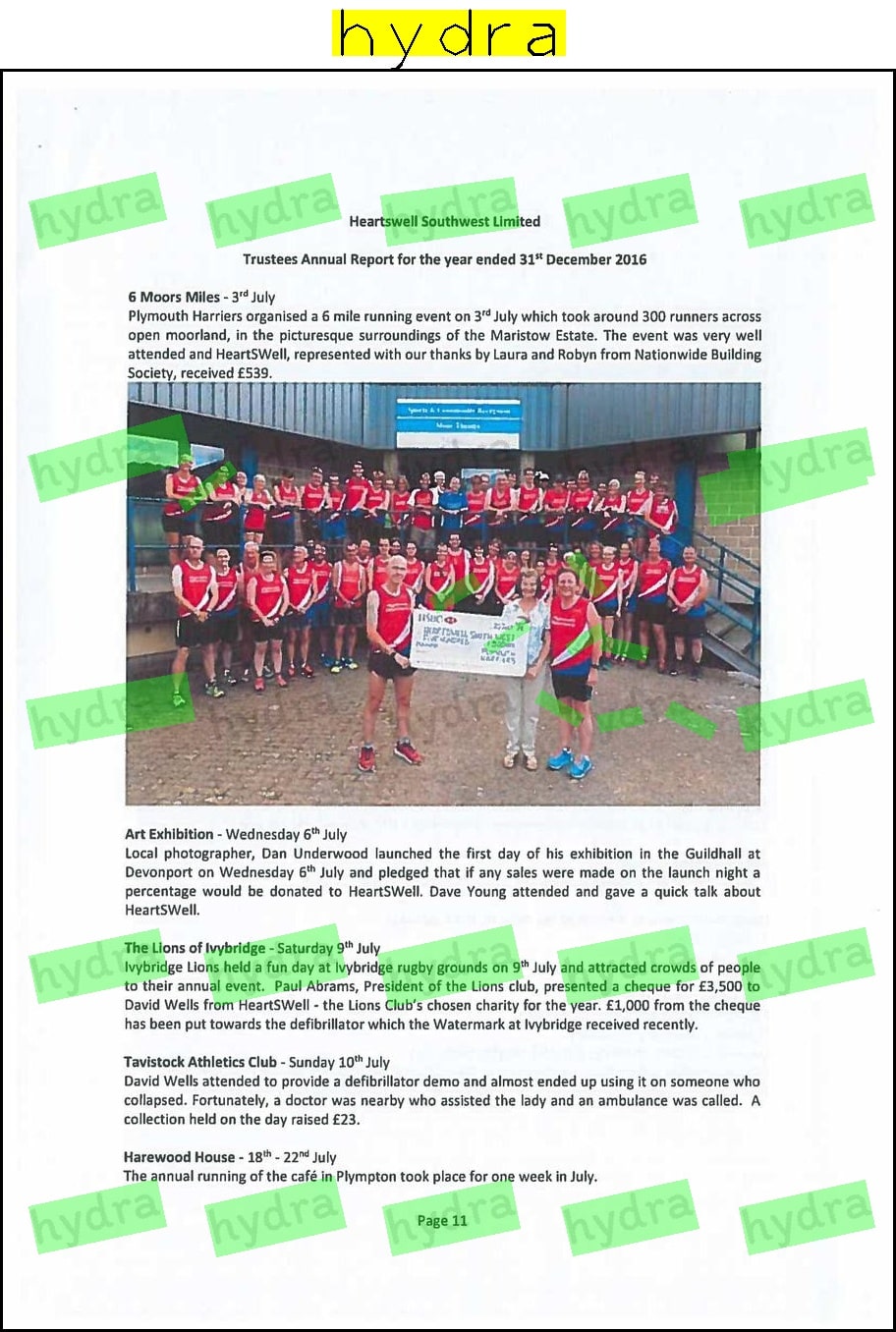} \\

		\end{tabular}}
	\end{center}
	\caption{\textbf{Watermark Text Detection Failure Cases of $\mathcal{W}$extract on K-Watermark.} Detections are highlighted with green bounding boxes and recognized text is written as yellow highlighted text at the top of each image. Failure situations usually occur due to overlap between non-uniform or dark-coloured visual elements, which impacts text recognition (\textit{e.g.}, leftmost image, \texttt{significance} vs. \texttt{significane}). These scenarios are complex, even for humans.}
	\label{fig:failures_detection}
\end{figure}

On top of the global and local feature representation $\mathrm{\Theta}$ we place an auto-regressive character-level transformer decoder \cite{vaswani_nips_2017}, $\tilde{s}_i = \mathrm{\Psi}_\texttt{TXT}(\mathrm{\Theta}, \{\tilde{s}_j\}_{1 \leq j \leq i})$, to generate the depicted watermark text sequence $\tilde{s}$.

\noindent In practice, the $i^{th}$ character $s_{i}$ is generated by the decoder $\mathrm{\Psi}_\texttt{TXT}$ by looking at the visual features $\mathrm{\Theta}$  and the previously generated characters (\textit{i.e.}, $\forall s_j$ with $j<i$).

During training, we use the reference watermark text $s$ and train with the cross-entropy loss $\mathcal{L}_{\texttt{TXT}}(\tilde{s}, s)$. During inference, we generate the watermark text, character by character, until the \texttt{[EOW]} symbol is generated or the maximal sequence length (in our experiments is set to $15$).
The idea behind our approach is to combine the information from the visual features (global and local) using the hierarchical encoder, with an auto-regressive decoding step that enables semantic reasoning over the previously generated characters. 

\begin{figure*} [p]
	\begin{center}
		\scalebox{0.77}{
			\setlength{\tabcolsep}{5pt}
			\begin{tabular}{ccccc}
				\includegraphics[width=0.2\linewidth]{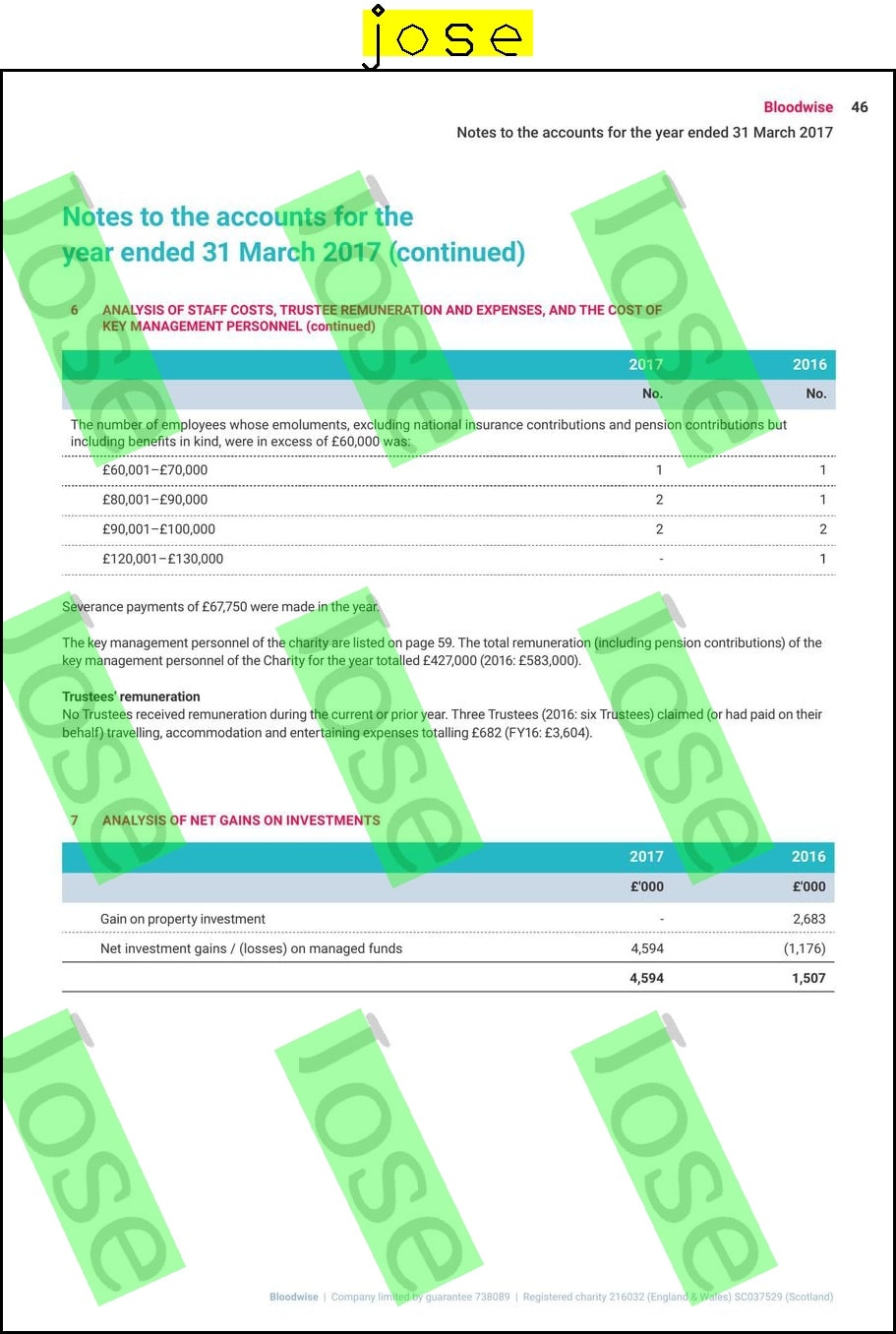} &
				\includegraphics[width=0.2\linewidth]{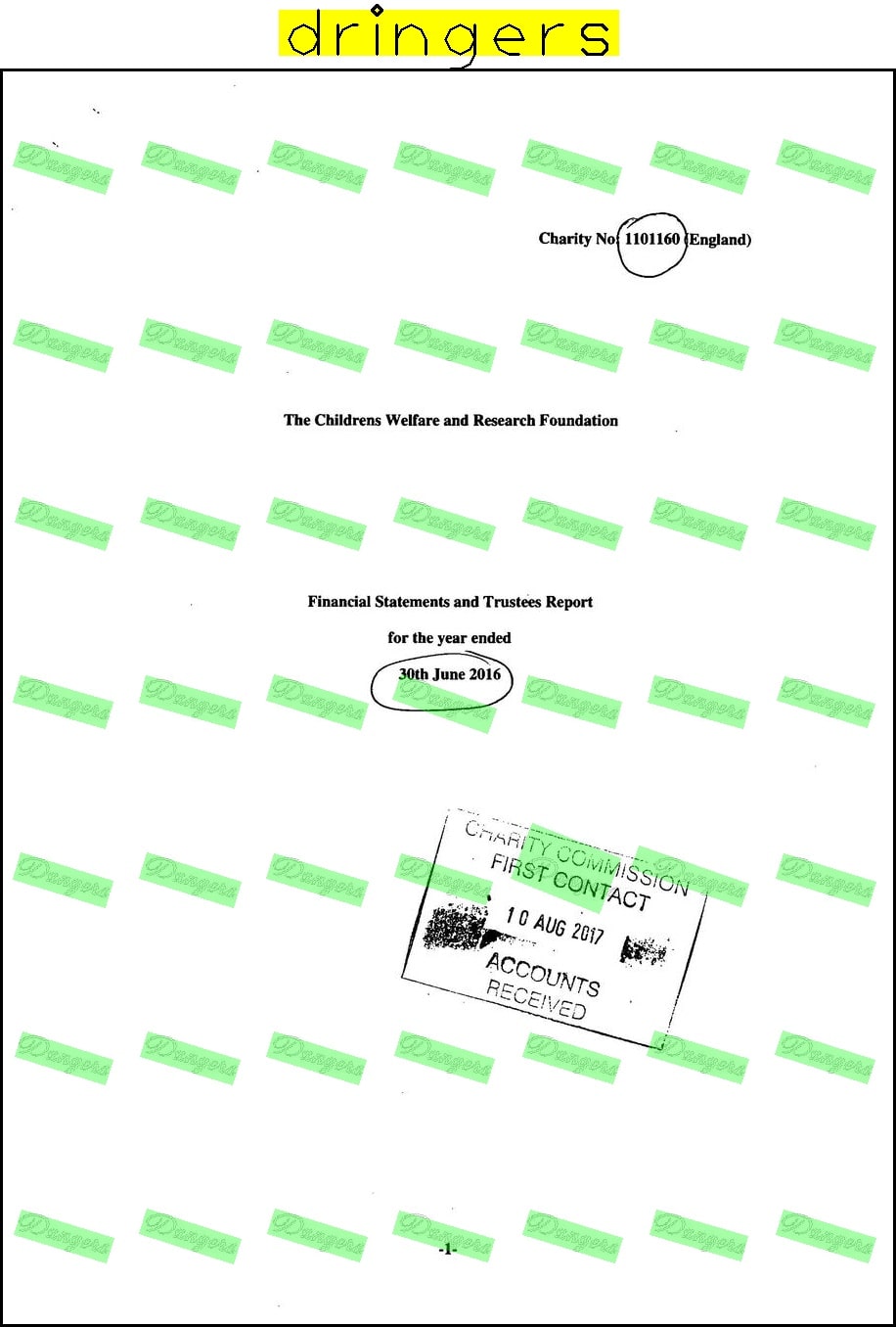} &
				\includegraphics[width=0.2\linewidth]{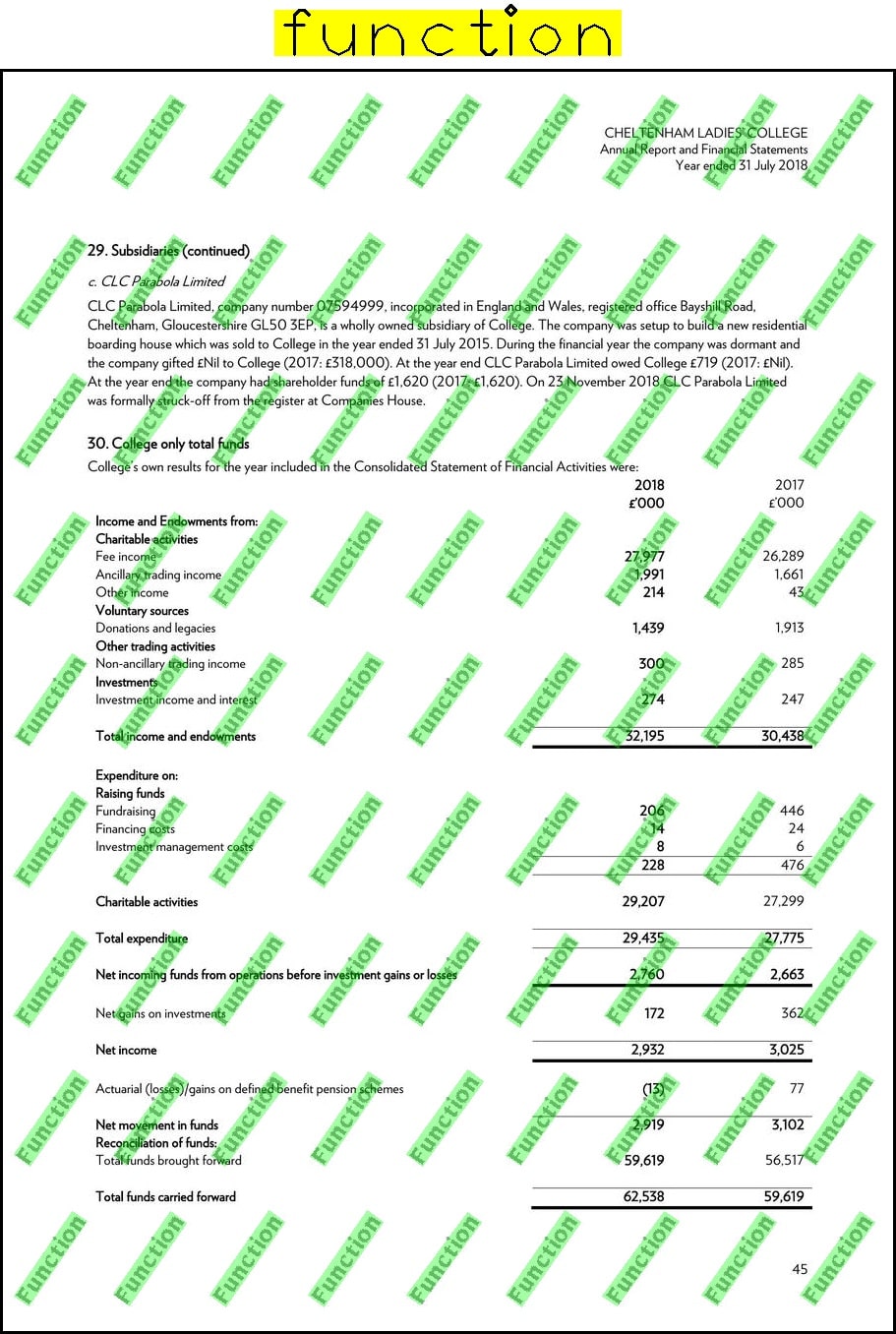} &
				\includegraphics[width=0.2\linewidth]{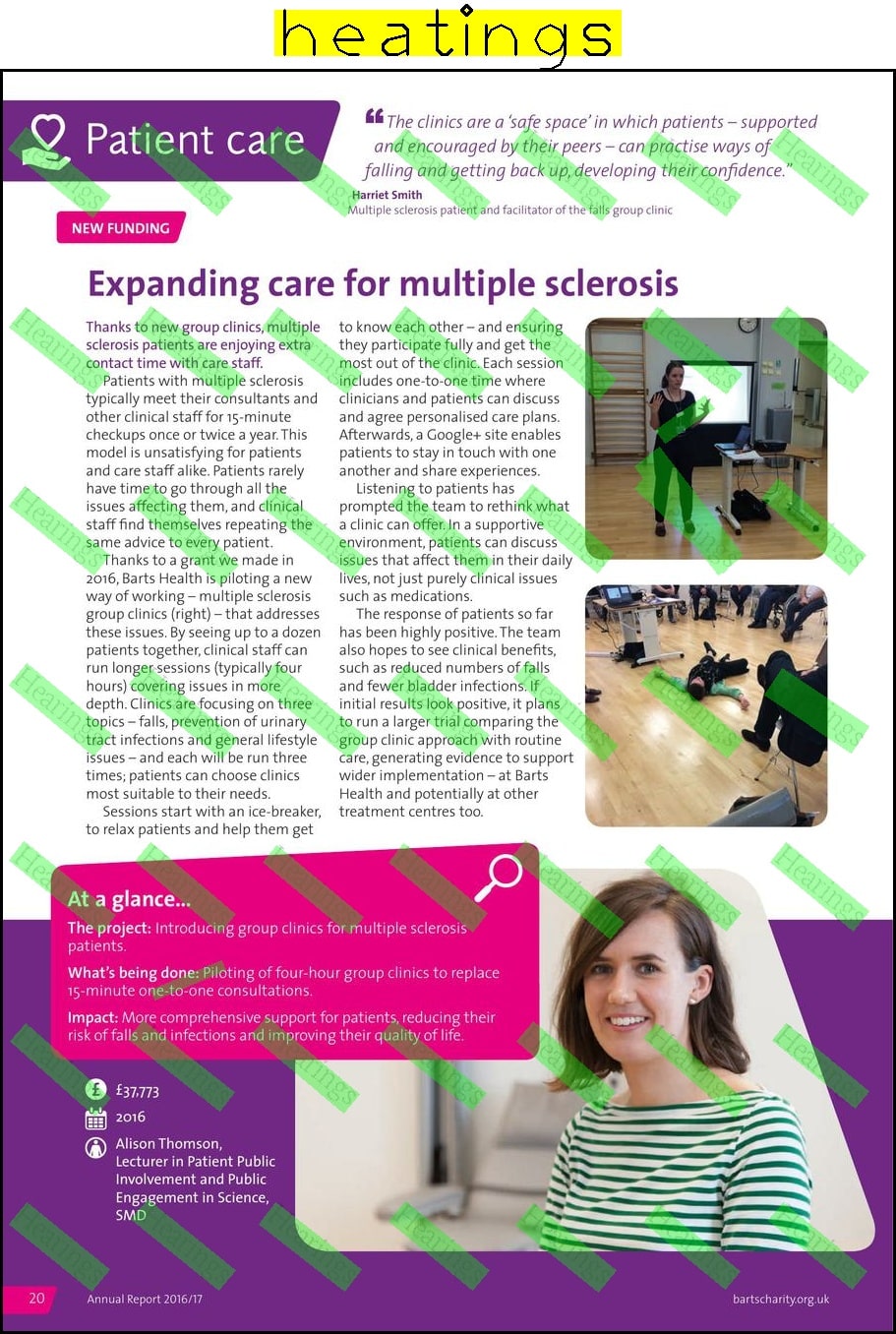} & \includegraphics[width=0.2\linewidth]{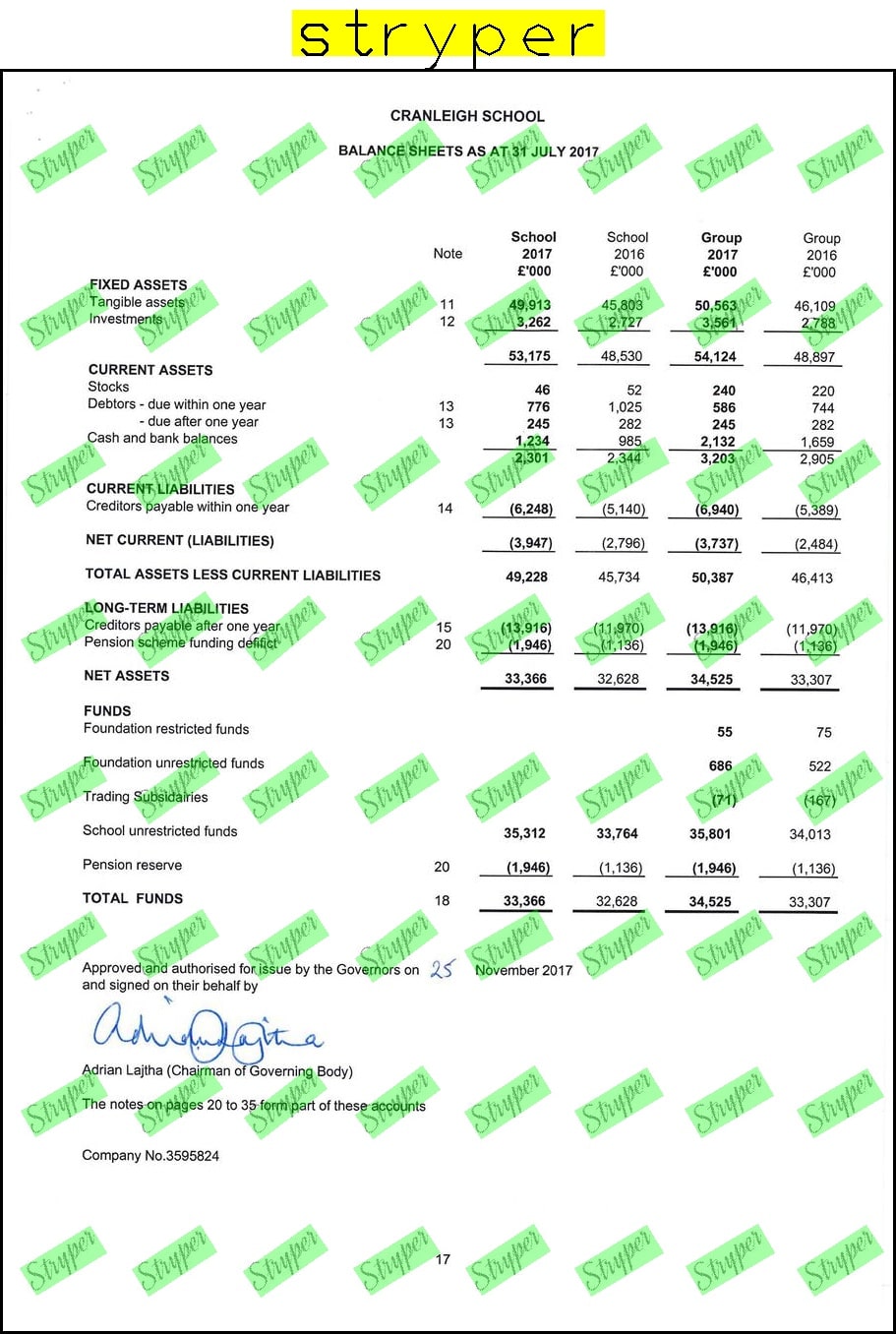}
				\\
				\includegraphics[width=0.2\linewidth,height=5.3cm]{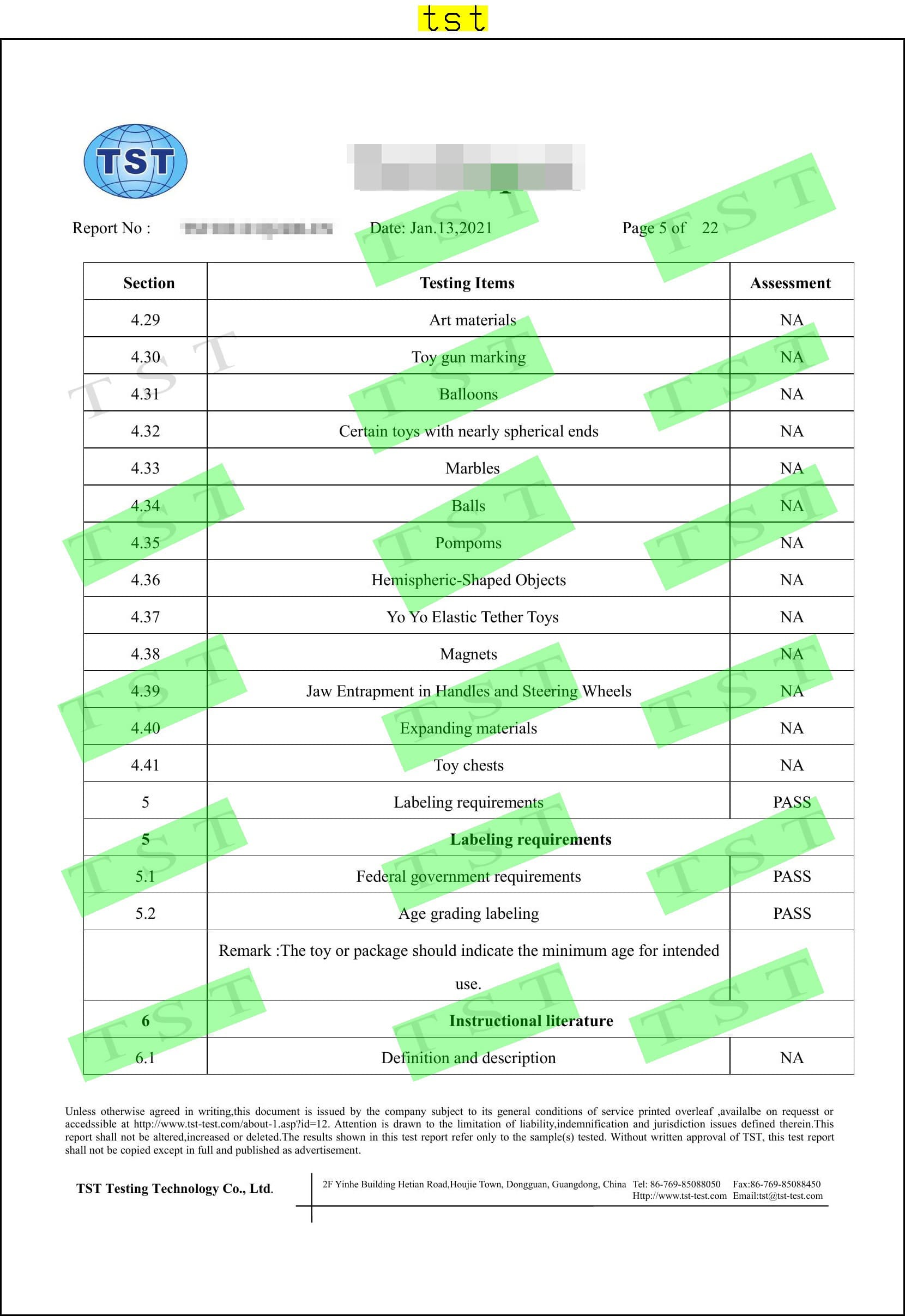} &
				\includegraphics[width=0.2\linewidth,height=5.5cm]{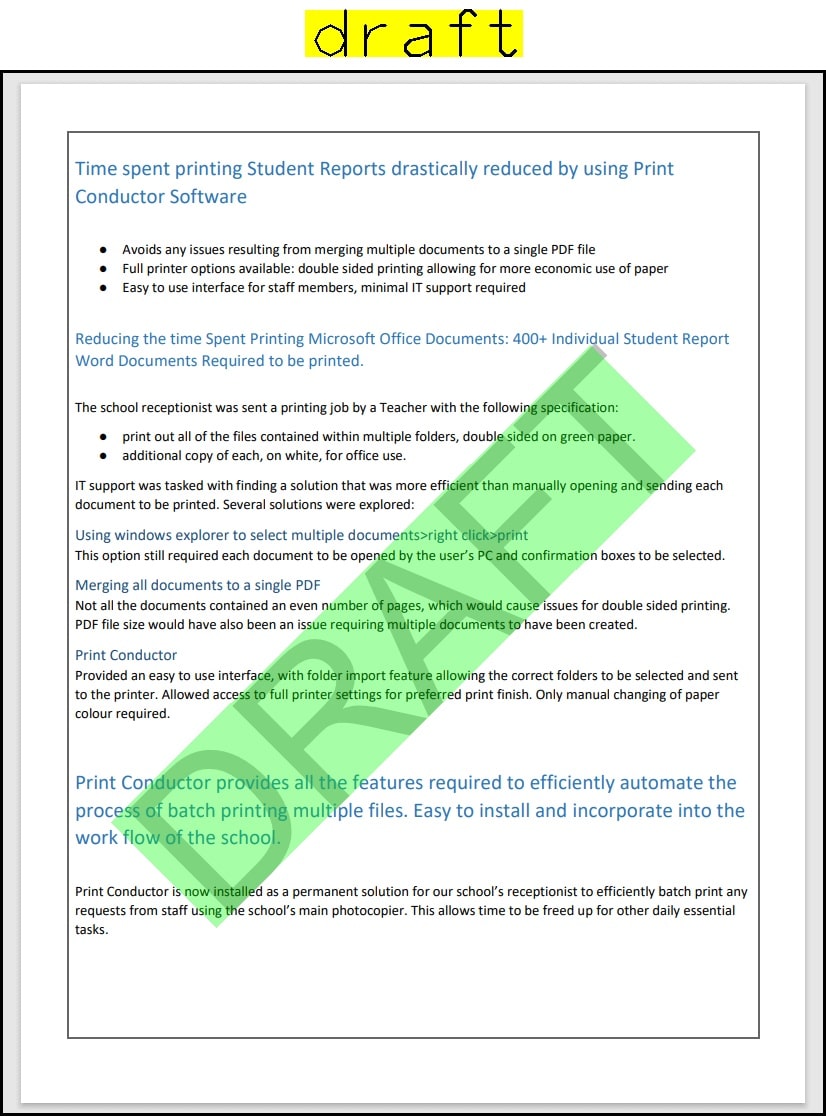} &
				\includegraphics[width=0.2\linewidth,height=5.3cm]{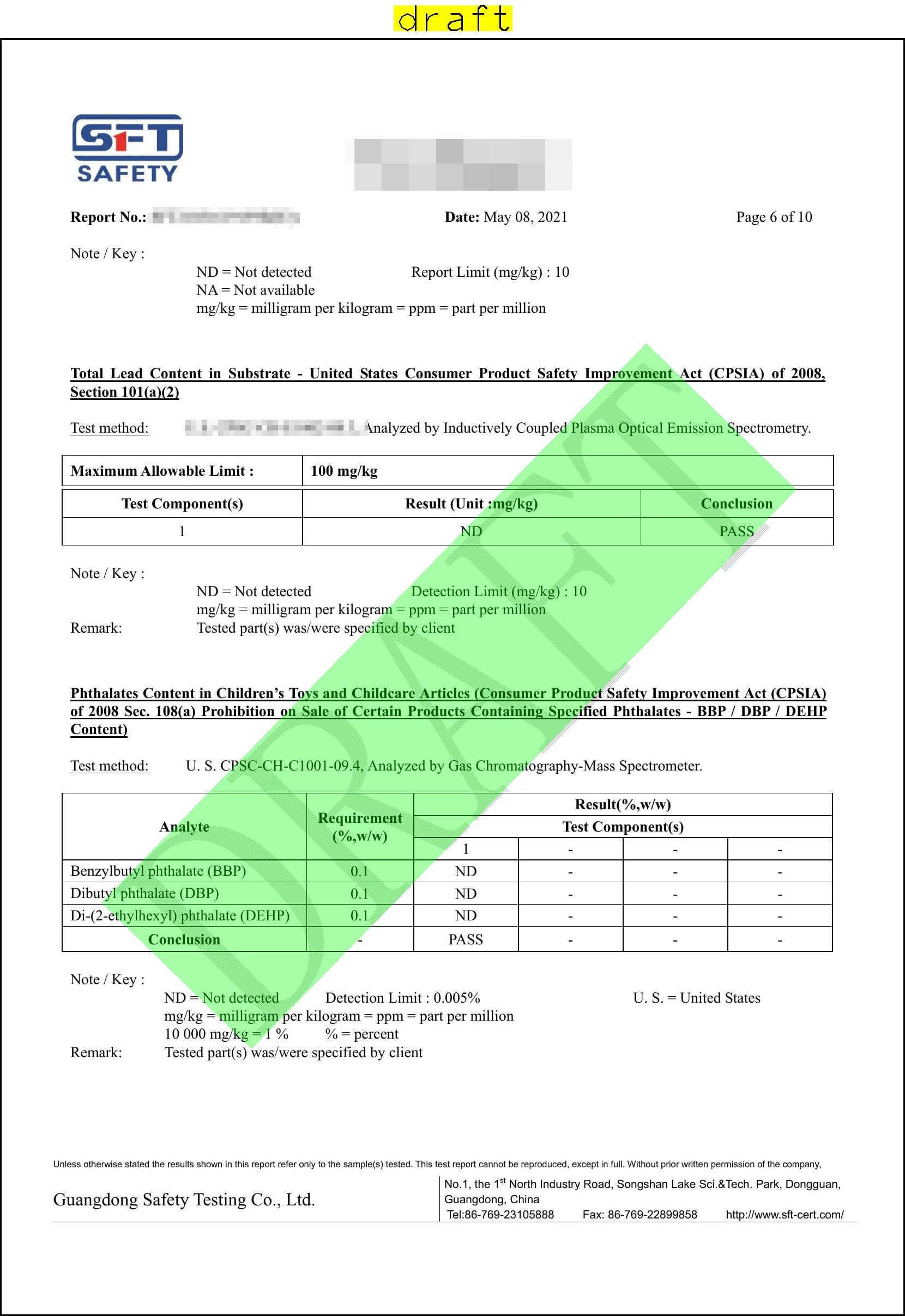} &
				\includegraphics[width=0.2\linewidth,height=5.4cm]{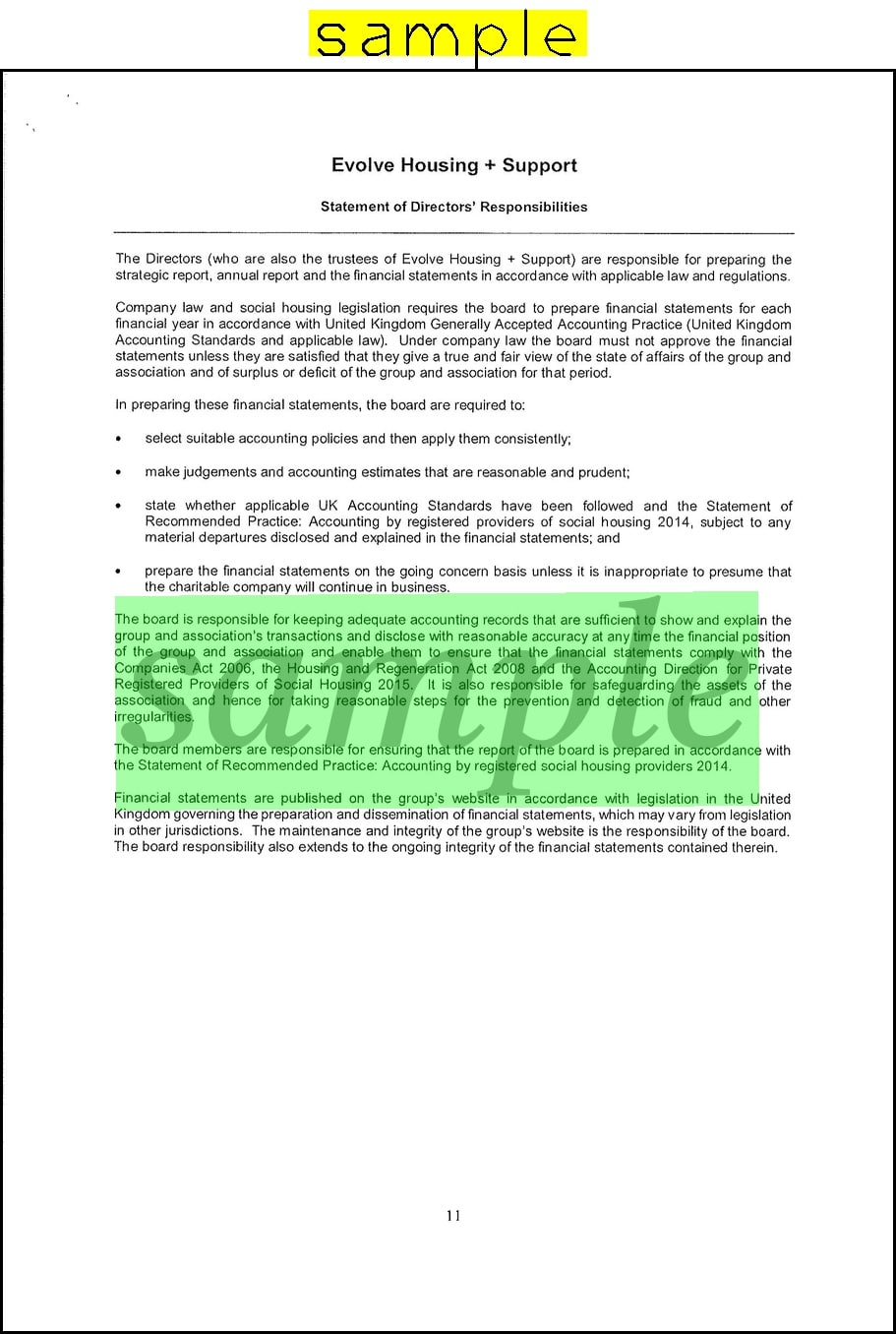} &
				\includegraphics[width=0.2\linewidth,height=5.45cm]{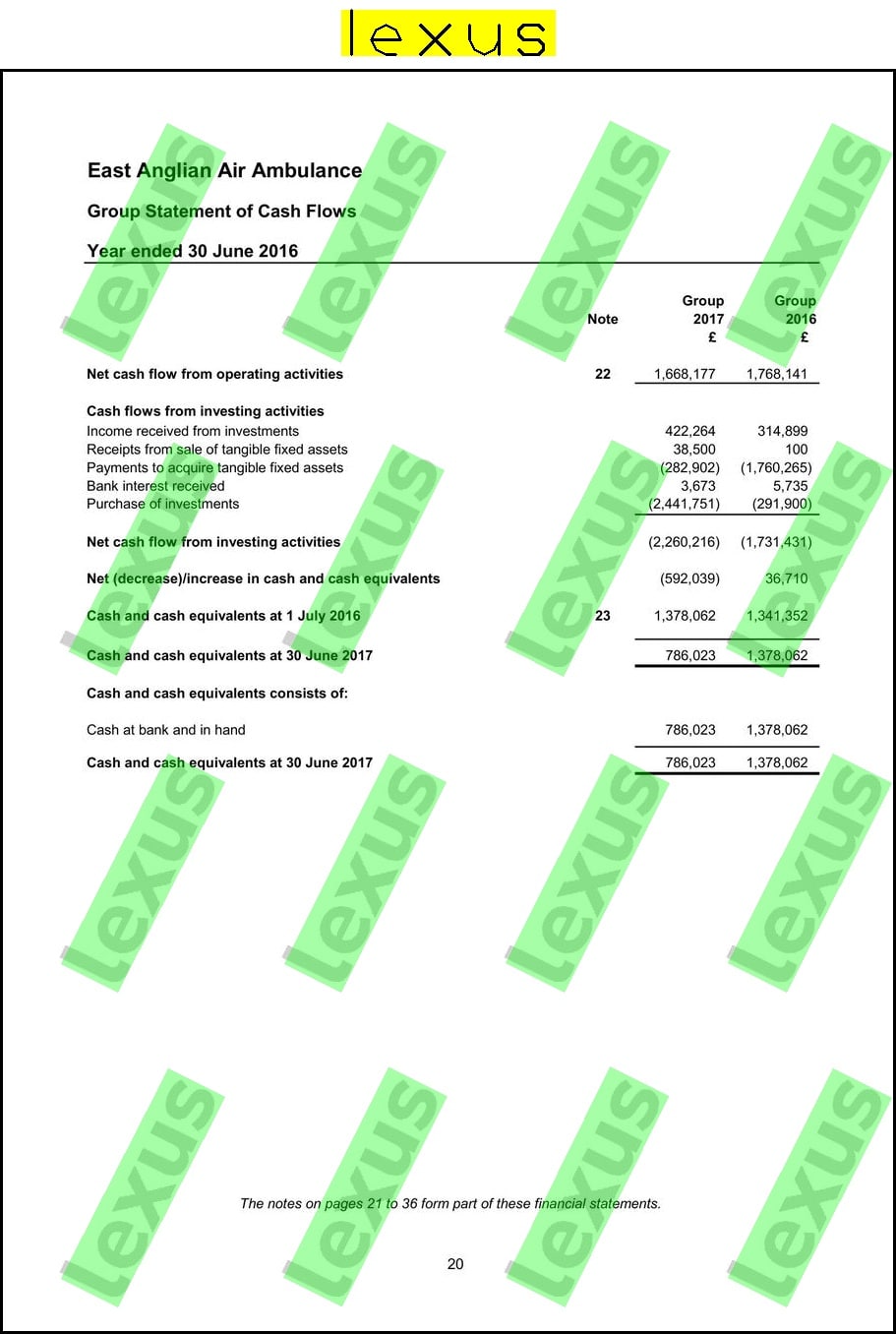}
				\\
				\includegraphics[width=0.2\linewidth,height=5.1cm]{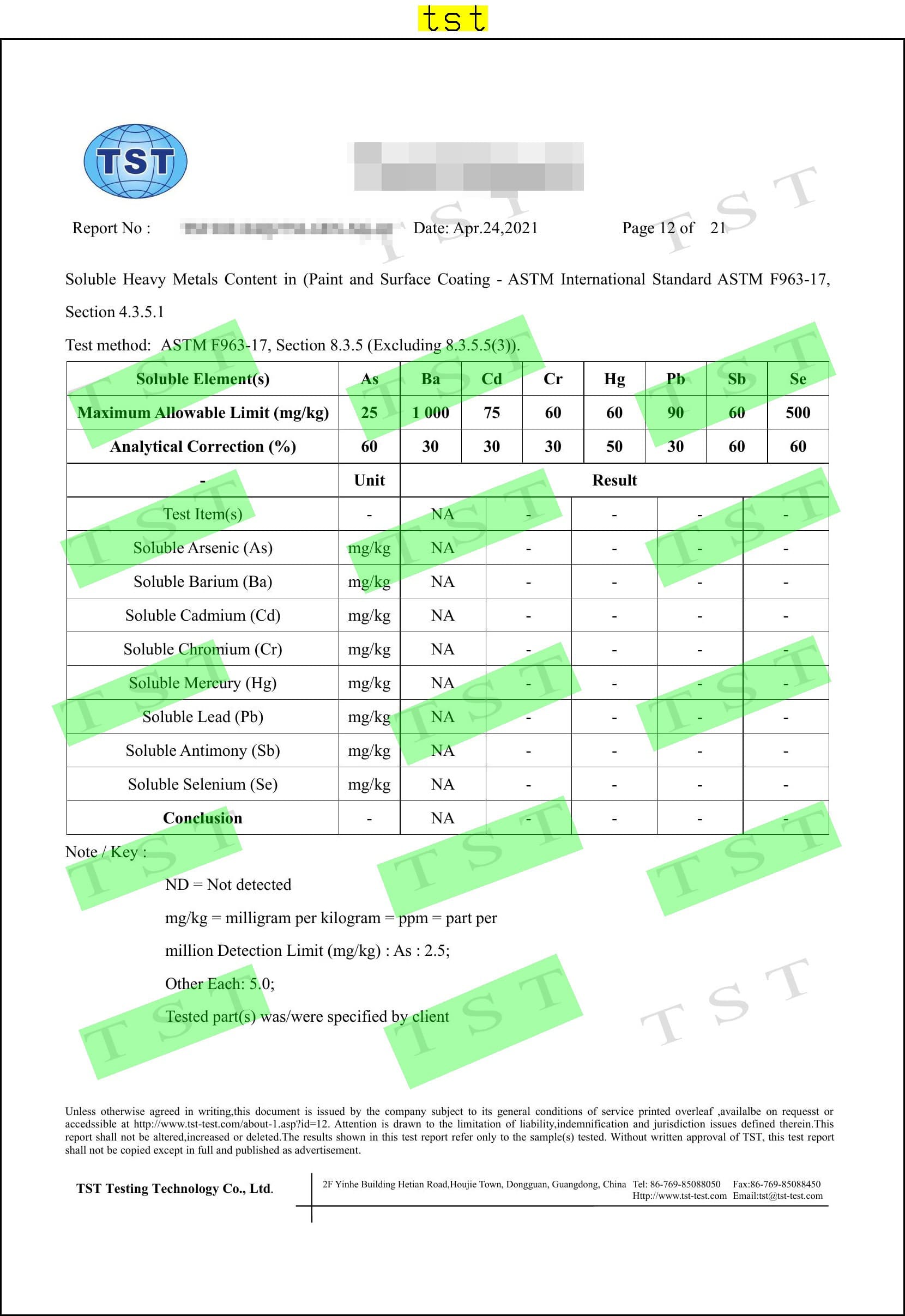} &
				\includegraphics[width=0.2\linewidth,height=5.5cm]{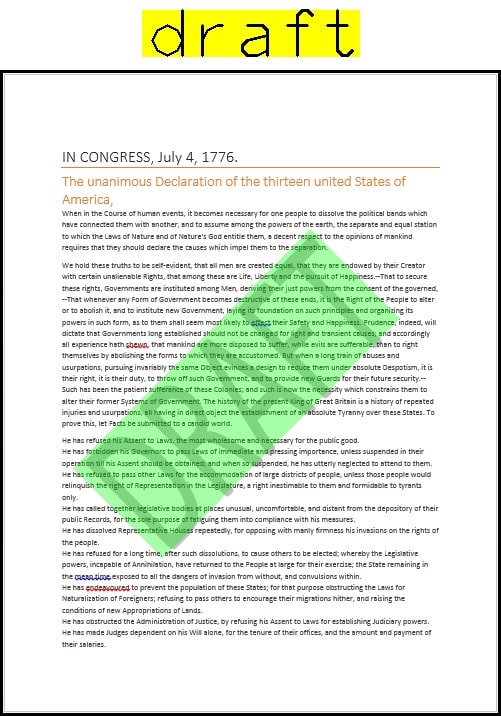} &
				\includegraphics[width=0.2\linewidth,height=5.2cm]{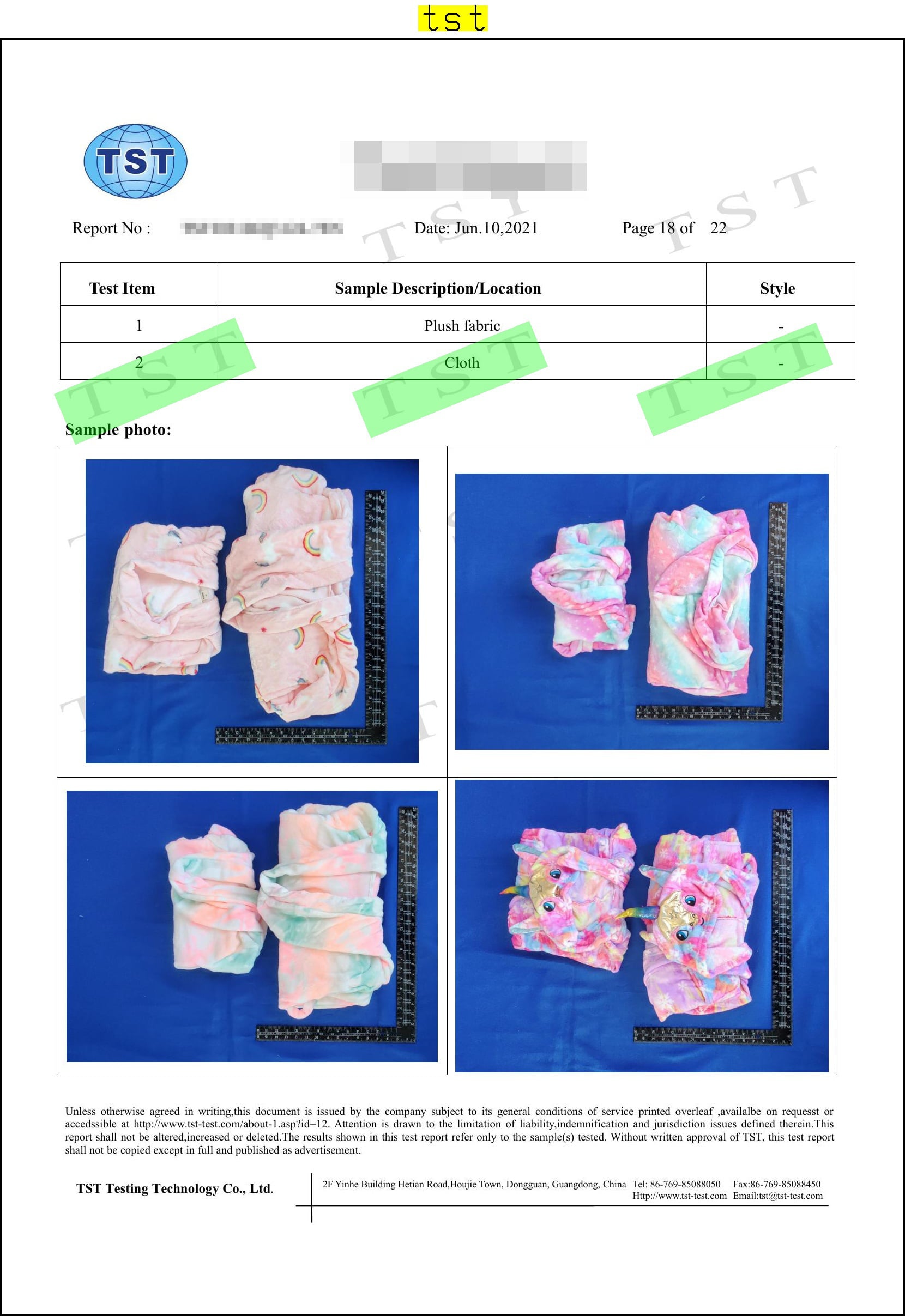} &
				\includegraphics[width=0.2\linewidth,height=5.2cm]{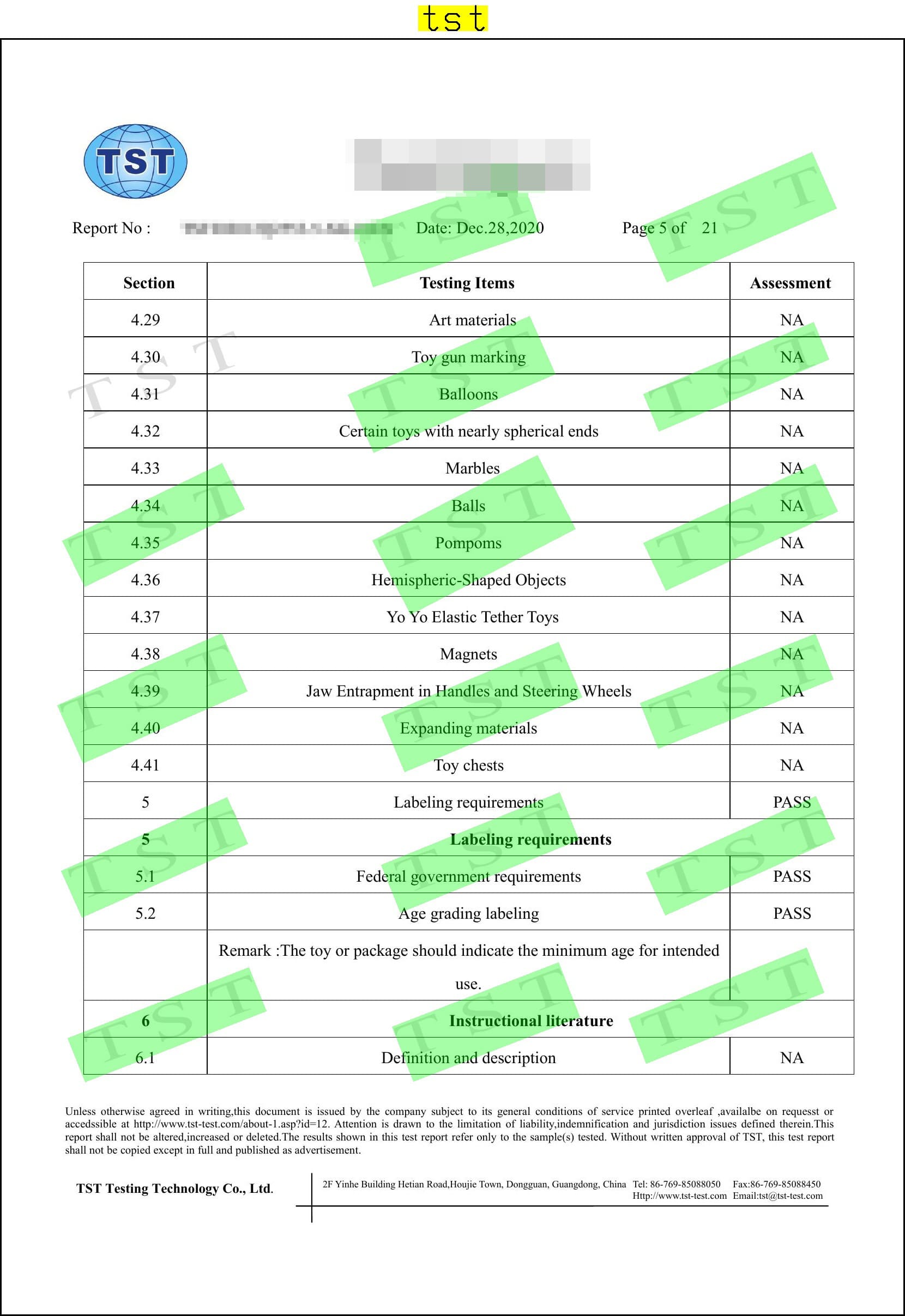} &
				\includegraphics[width=0.2\linewidth,height=5.2cm]{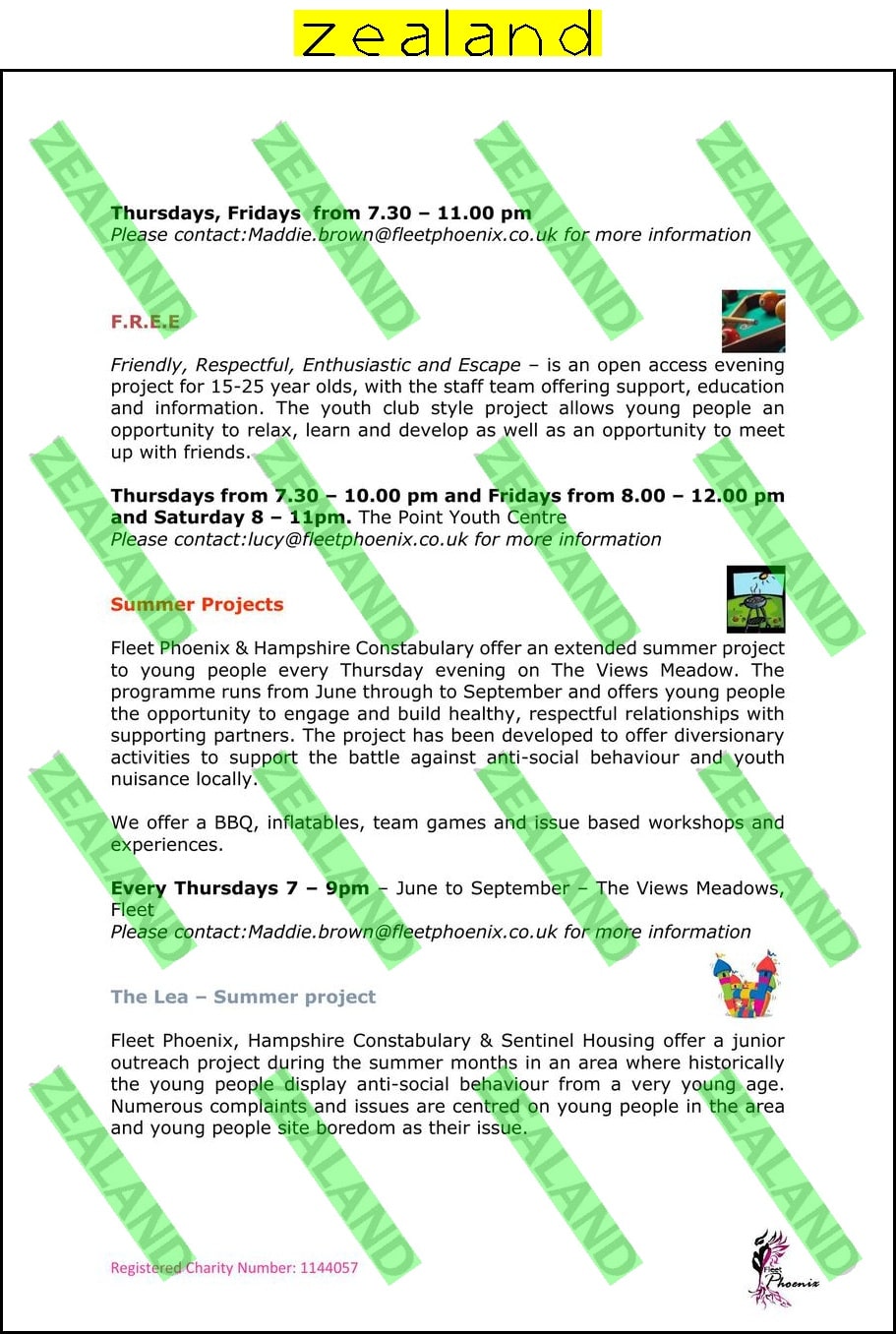}
		\end{tabular}}
	\end{center}
	\caption{\textbf{Visual Results of $\mathcal{W}$extract.} Row $(1)$ contains K-Watermark test samples and rows $(2)$ and $(3)$ contain web retrieved documents. Detections are highlighted with green bounding boxes and recognized text is written as yellow highlighted text at the top of the document page. Our approach exhibits robustness with respect to the density of the watermark text, the overlap degree with background text as well as images and the fadedness of the text. Additionally, it is robust with respect to vertical text orientation as well as other visual elements with diagonal orientation and generates consistent bounding box predictions in terms of angular orientation, width and height.}
	\label{fig:visual_result_b}
\end{figure*}

\begin{figure*} [!h]
	\centering
	\setlength{\tabcolsep}{5pt}
	\begin{tabular}{cc}
		\includegraphics[width=0.47\linewidth]{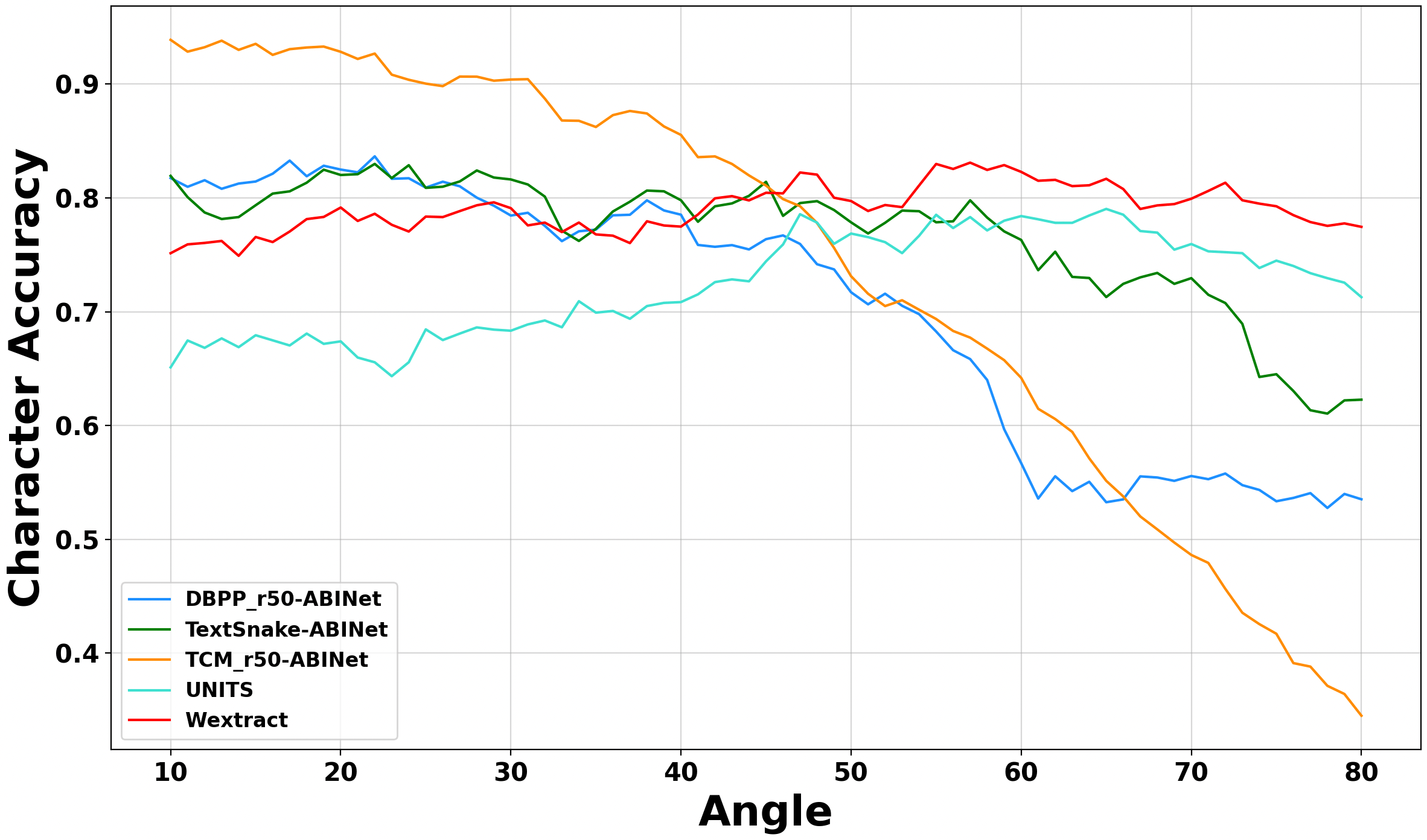} &
		\includegraphics[width=0.47\linewidth]{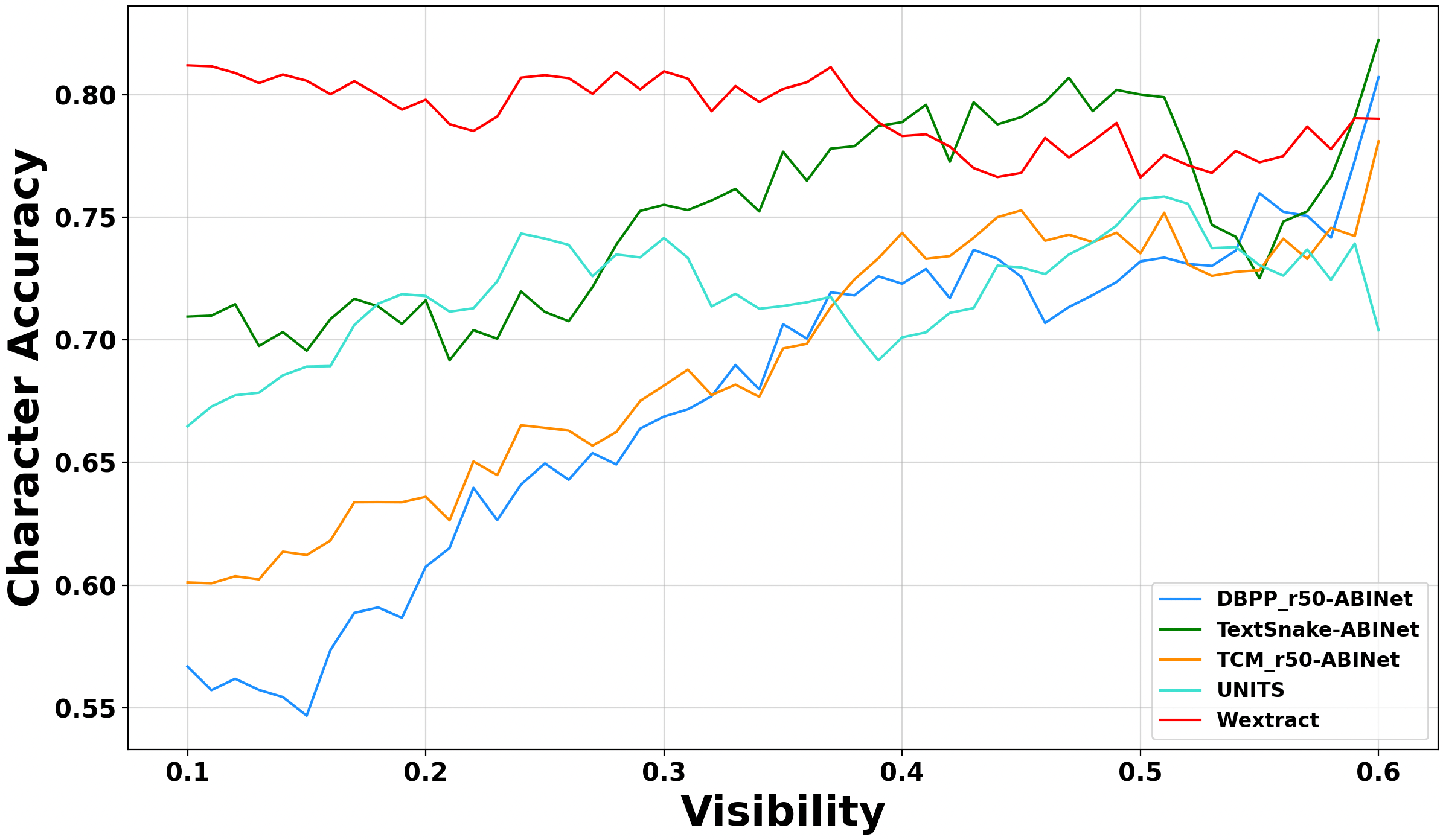}
		\end{tabular}
	\caption{\textbf{Watermark text spotting error plot against visibility and angular orientation (\textit{best seen in color}).} We plot the character accuracy metric for the \textbf{K-Watermark} dataset against different angular orientation of the watermark text with respect to the horizontal axis (\emph{left}) and different text visibility values (\emph{right}). Our method is robust with respect to the visibility and the angular orientation of the watermark text patterns and it is still being able to achieve a high performance even with near vertical angle or extreme fadedness.}
	\label{fig:plots_recognition}
\end{figure*}

\begin{figure*} [p]
	\begin{center}
		\scalebox{0.77}{
			\setlength{\tabcolsep}{5pt}
			\begin{tabular}{cccc}
			    (a) Original & (b) TCM \cite{yu2023turning}-ABInet\cite{fang2021read} & (c) UNITS \cite{kil2023towards} & (d) $\mathcal{W}$extract
			    \\
			    \includegraphics[width=0.22\linewidth]{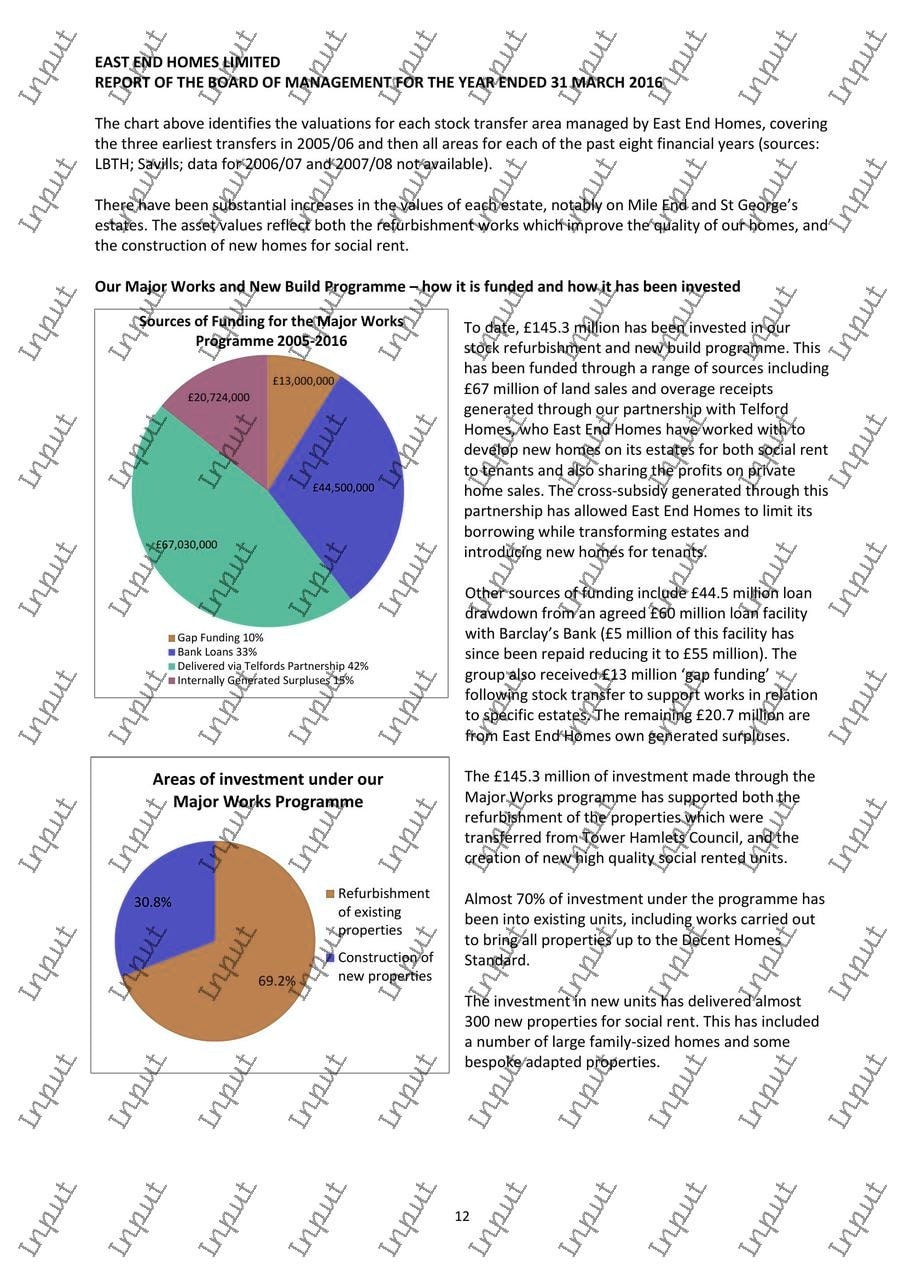} &
				\includegraphics[width=0.22\linewidth]{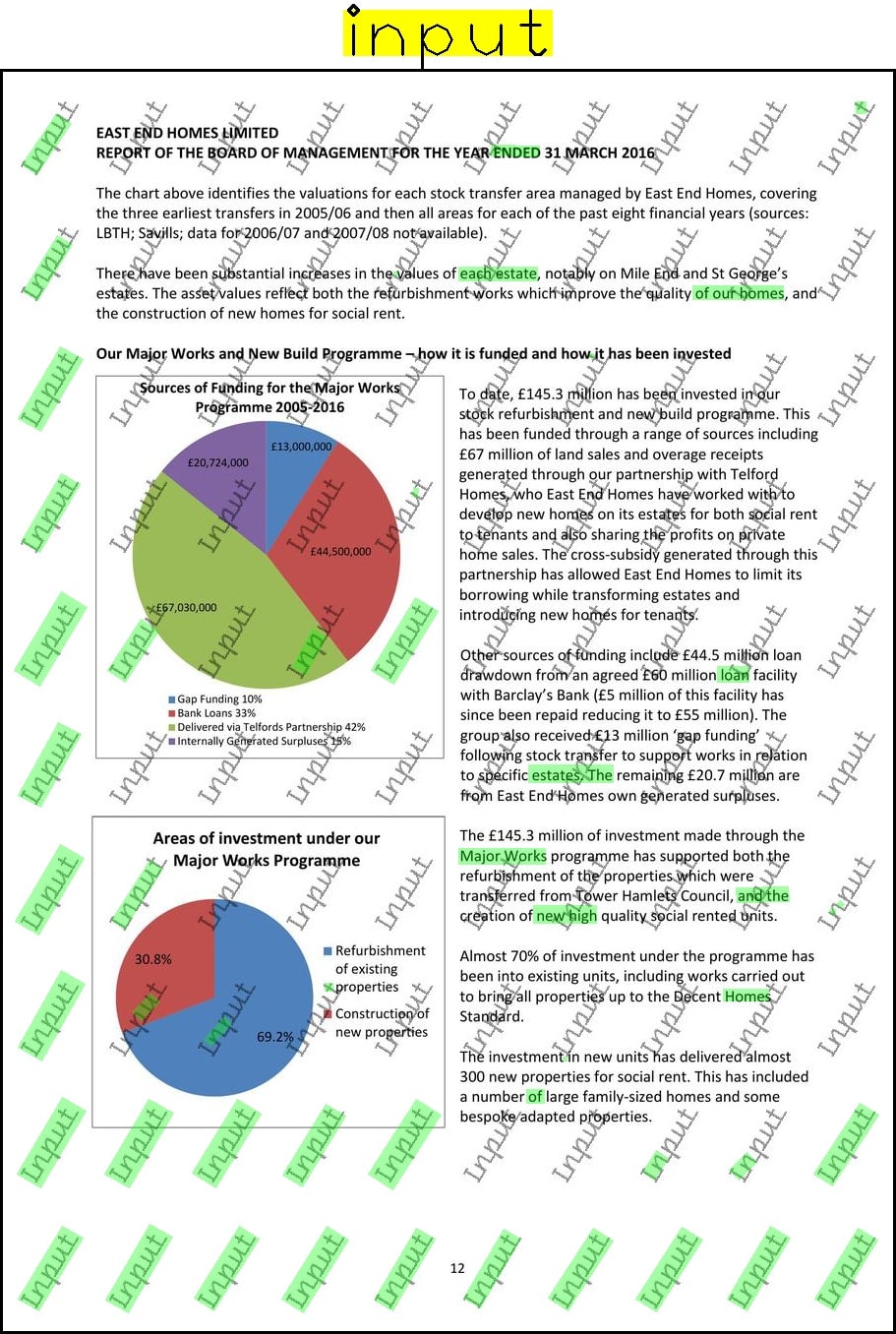} &
				\includegraphics[width=0.22\linewidth]{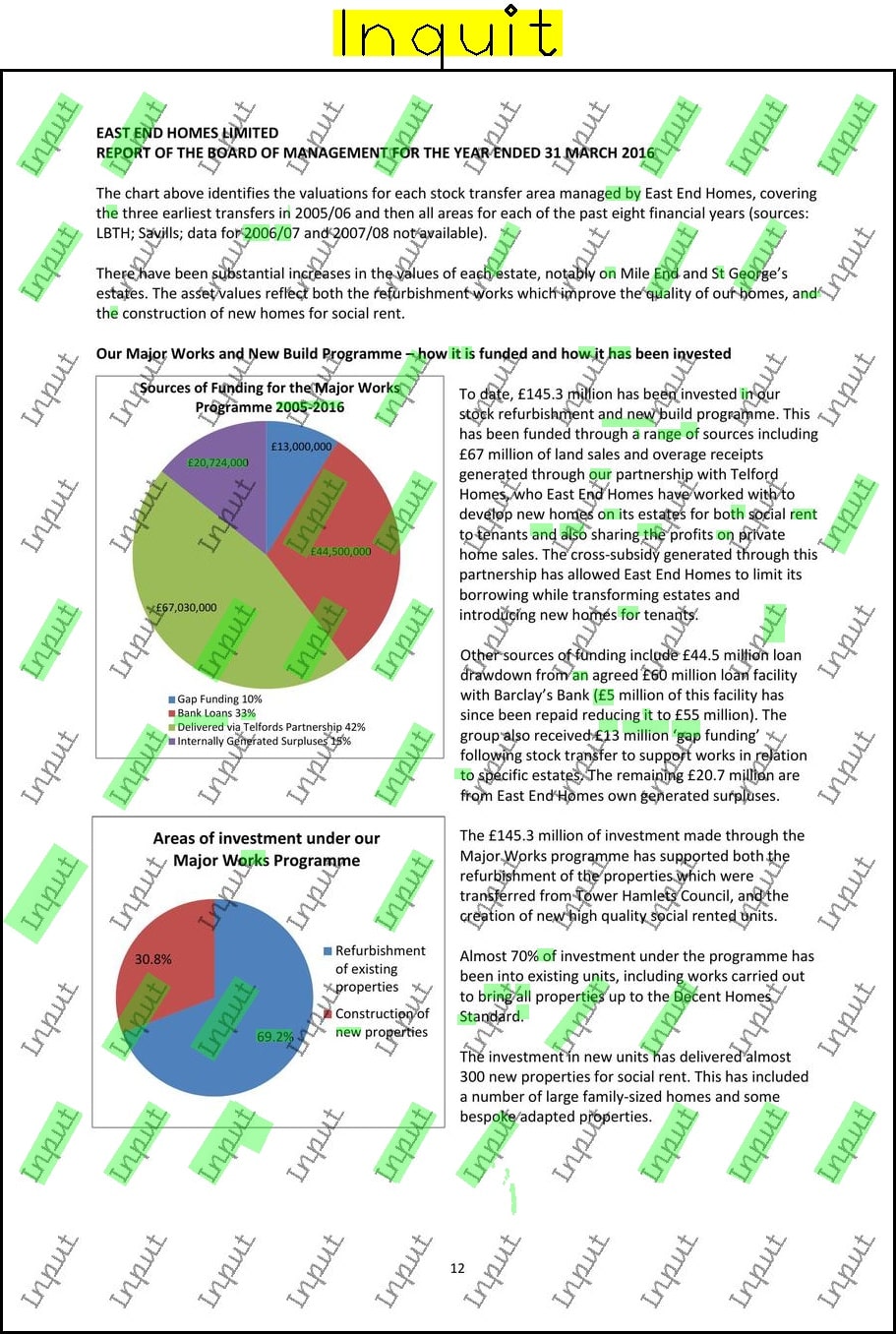} &
				\includegraphics[width=0.22\linewidth]{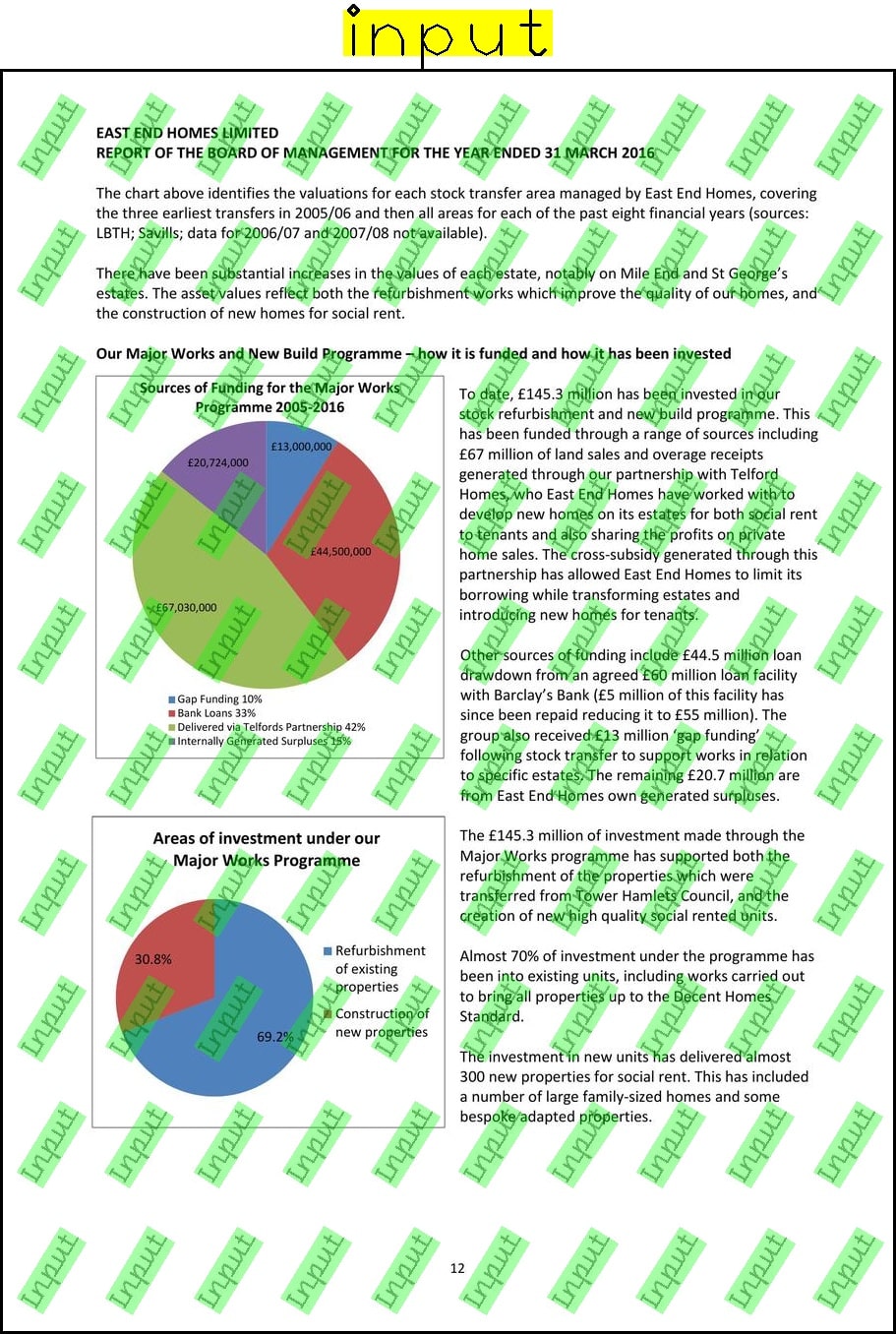}
				\\
				\includegraphics[width=0.2\linewidth]{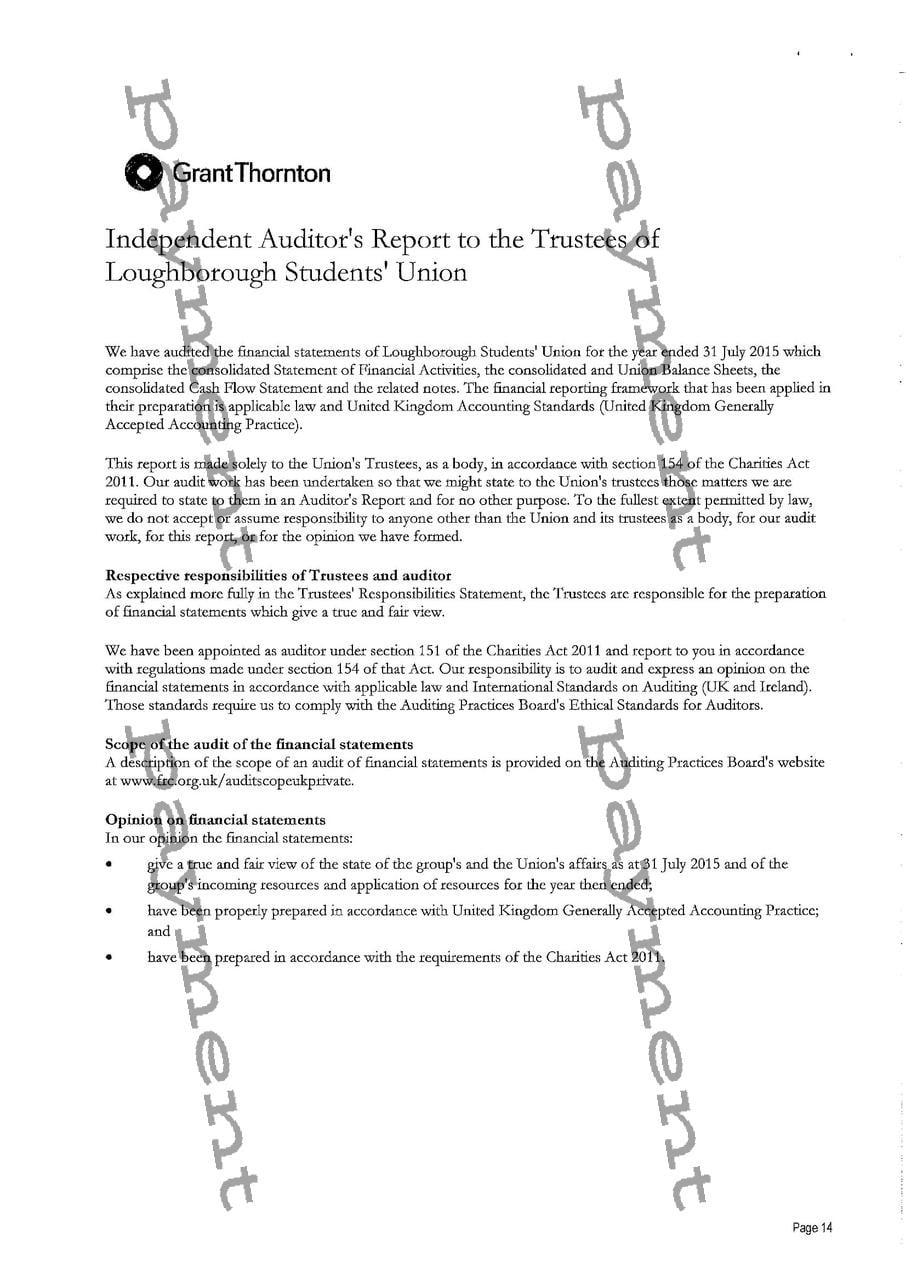} &
				\includegraphics[width=0.2\linewidth]{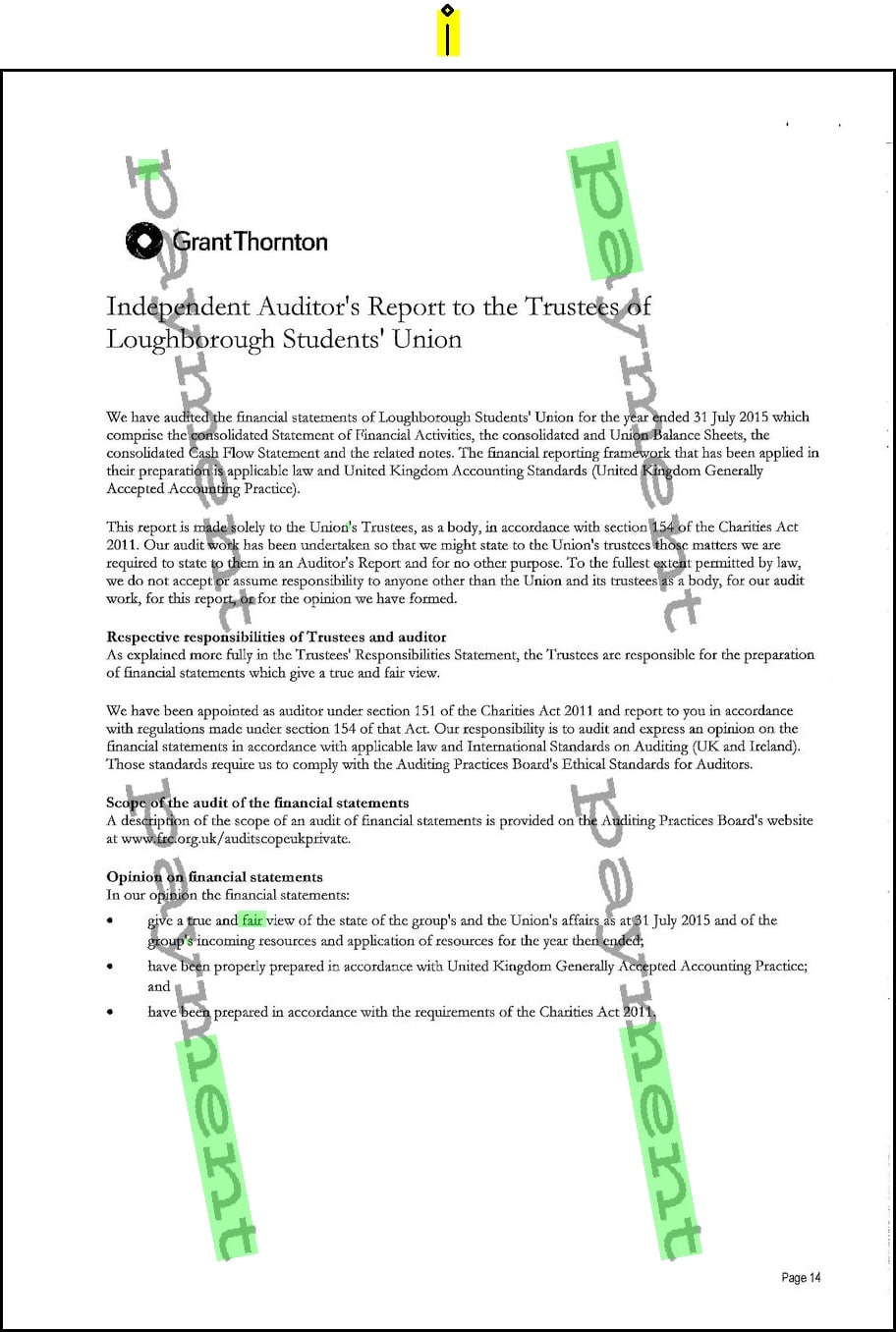} &
				\includegraphics[width=0.2\linewidth]{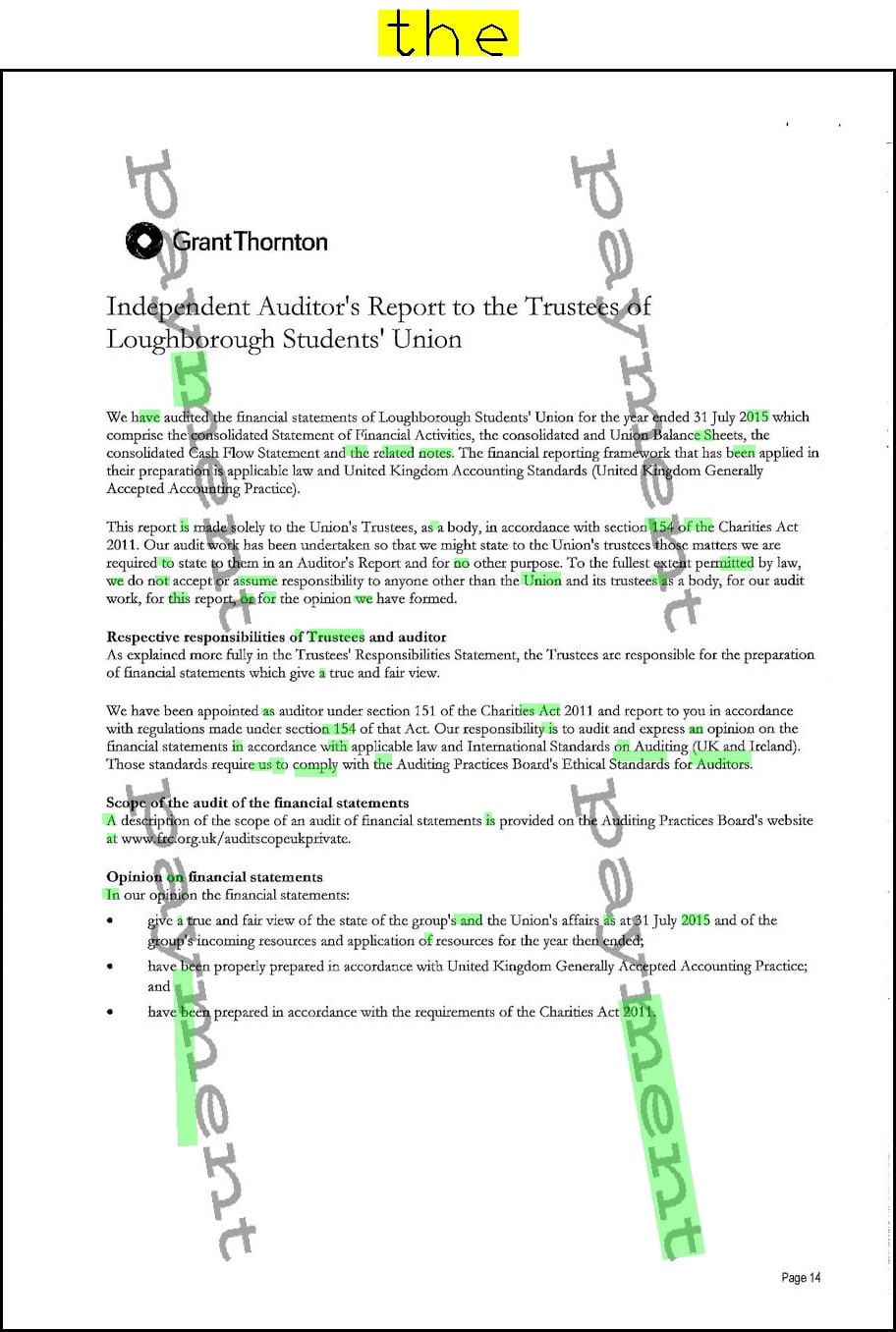} &
				\includegraphics[width=0.2\linewidth]{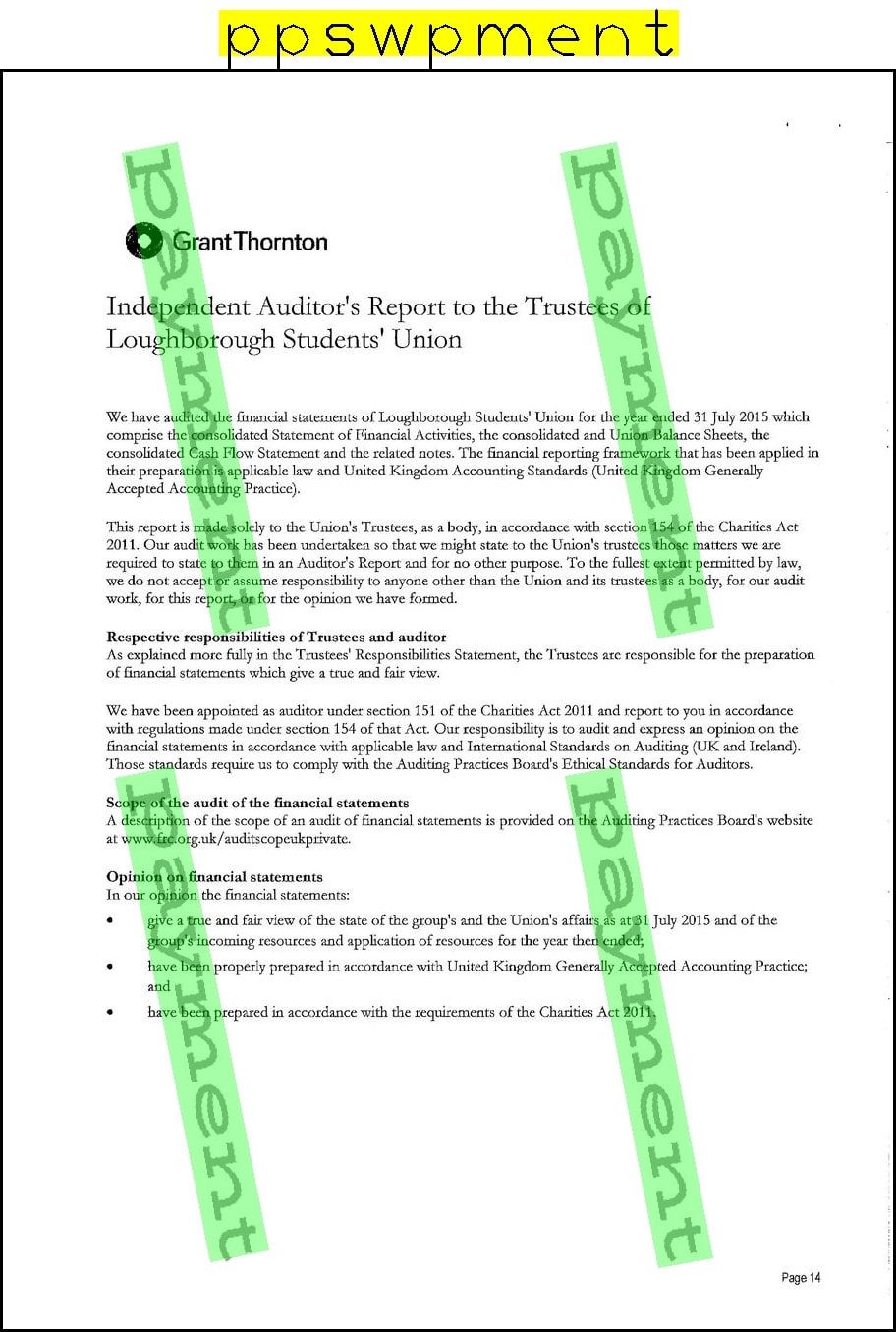}
				\\
		        \includegraphics[width=0.2\linewidth]{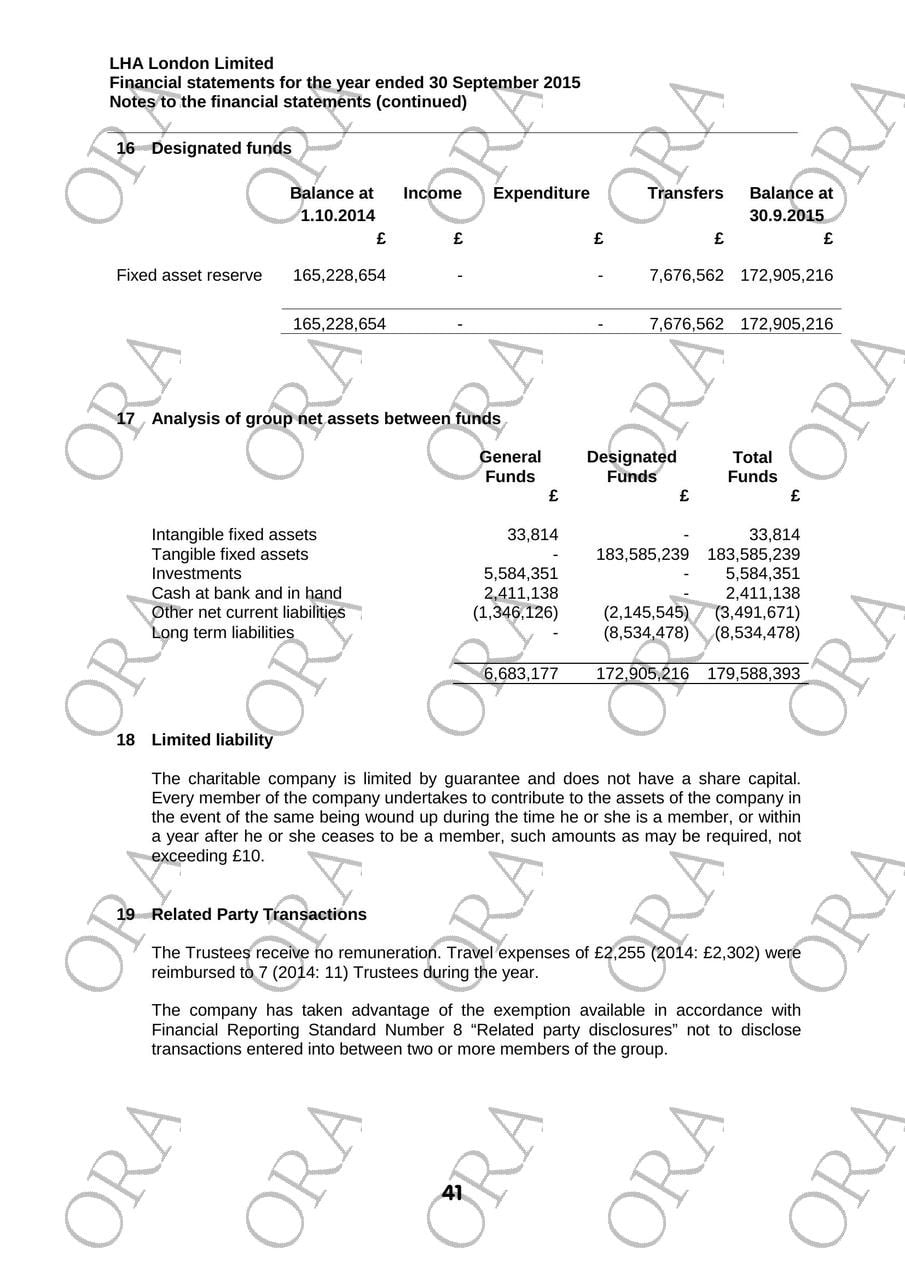} &
				\includegraphics[width=0.2\linewidth]{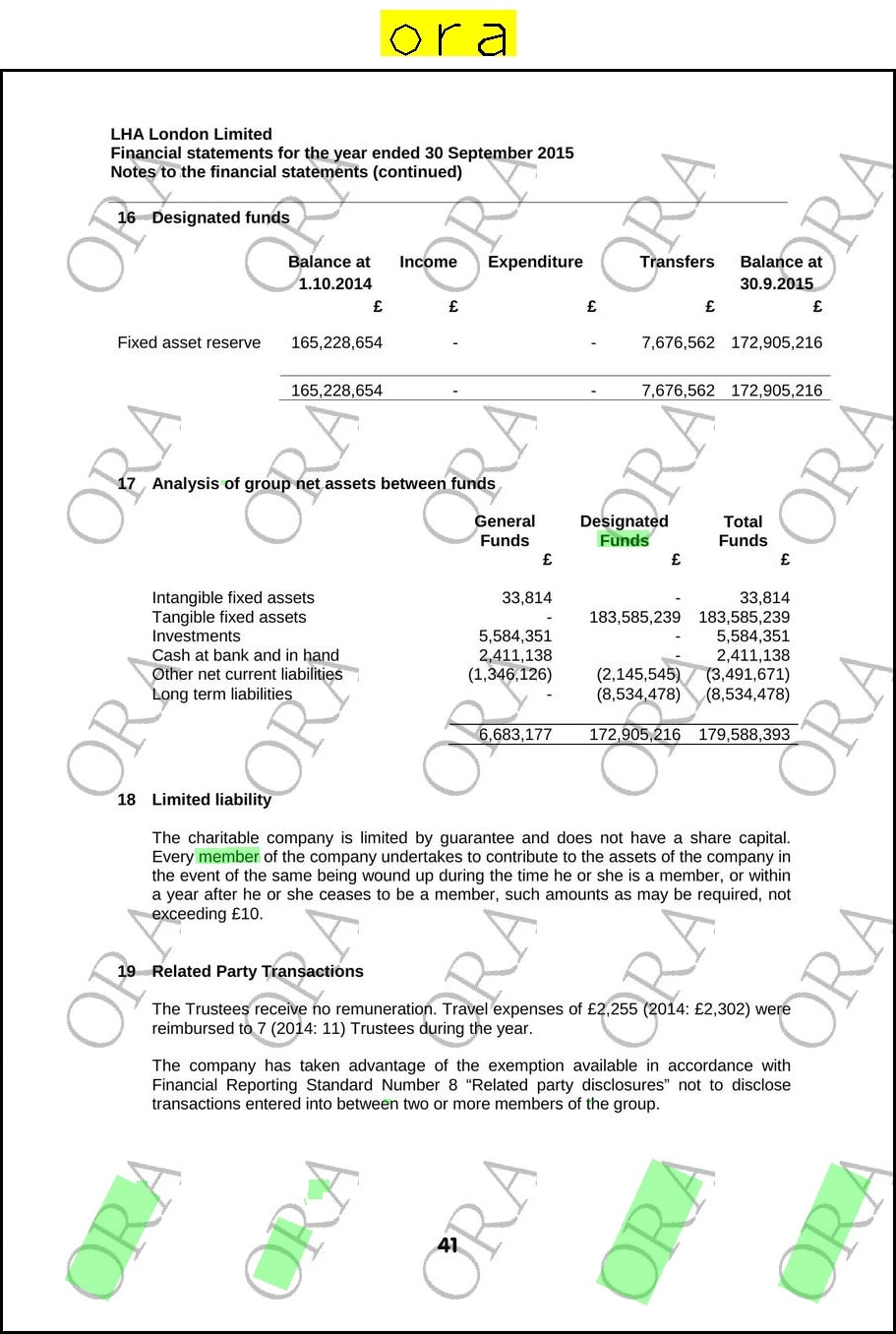} &
				\includegraphics[width=0.2\linewidth]{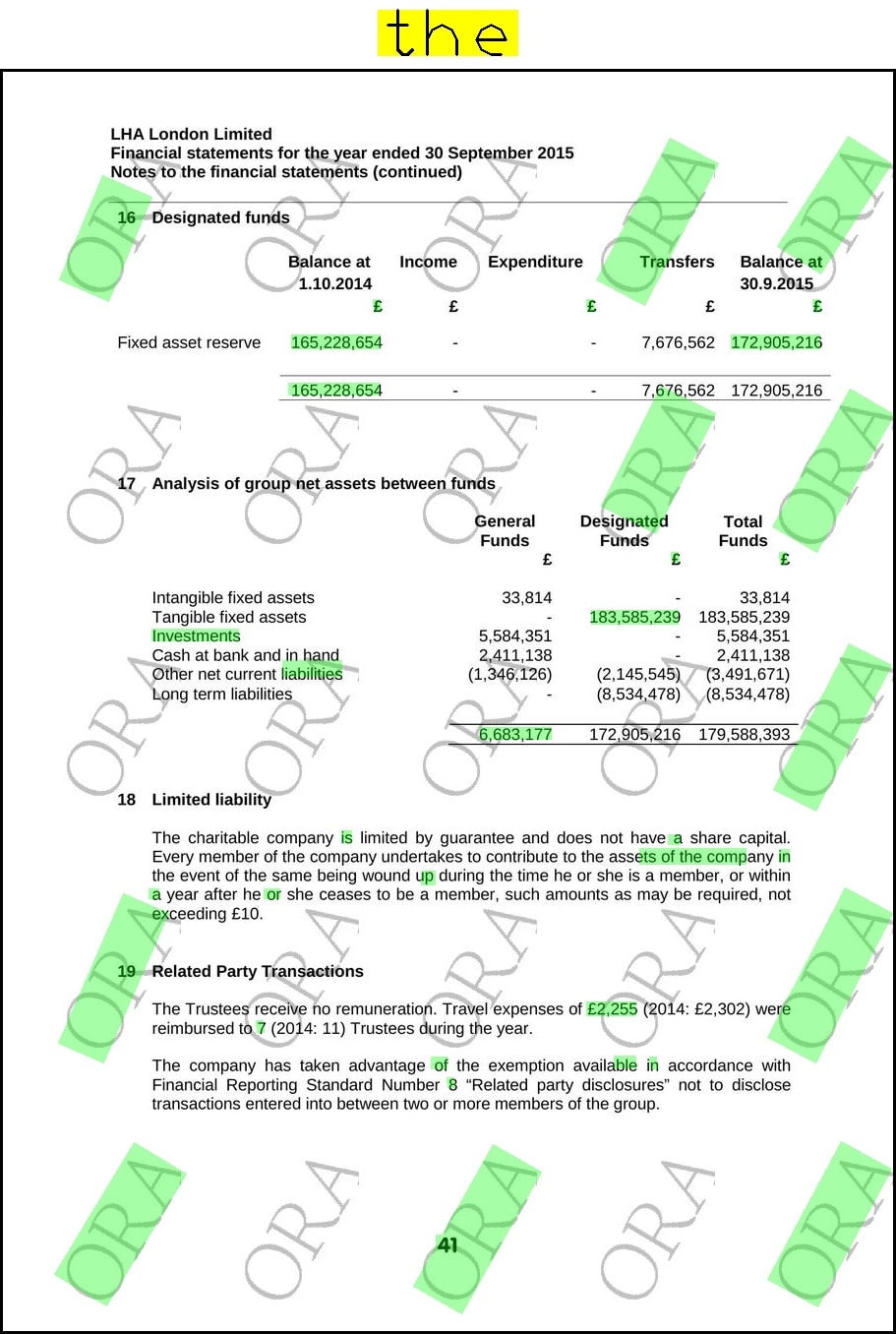} &
				\includegraphics[width=0.2\linewidth]{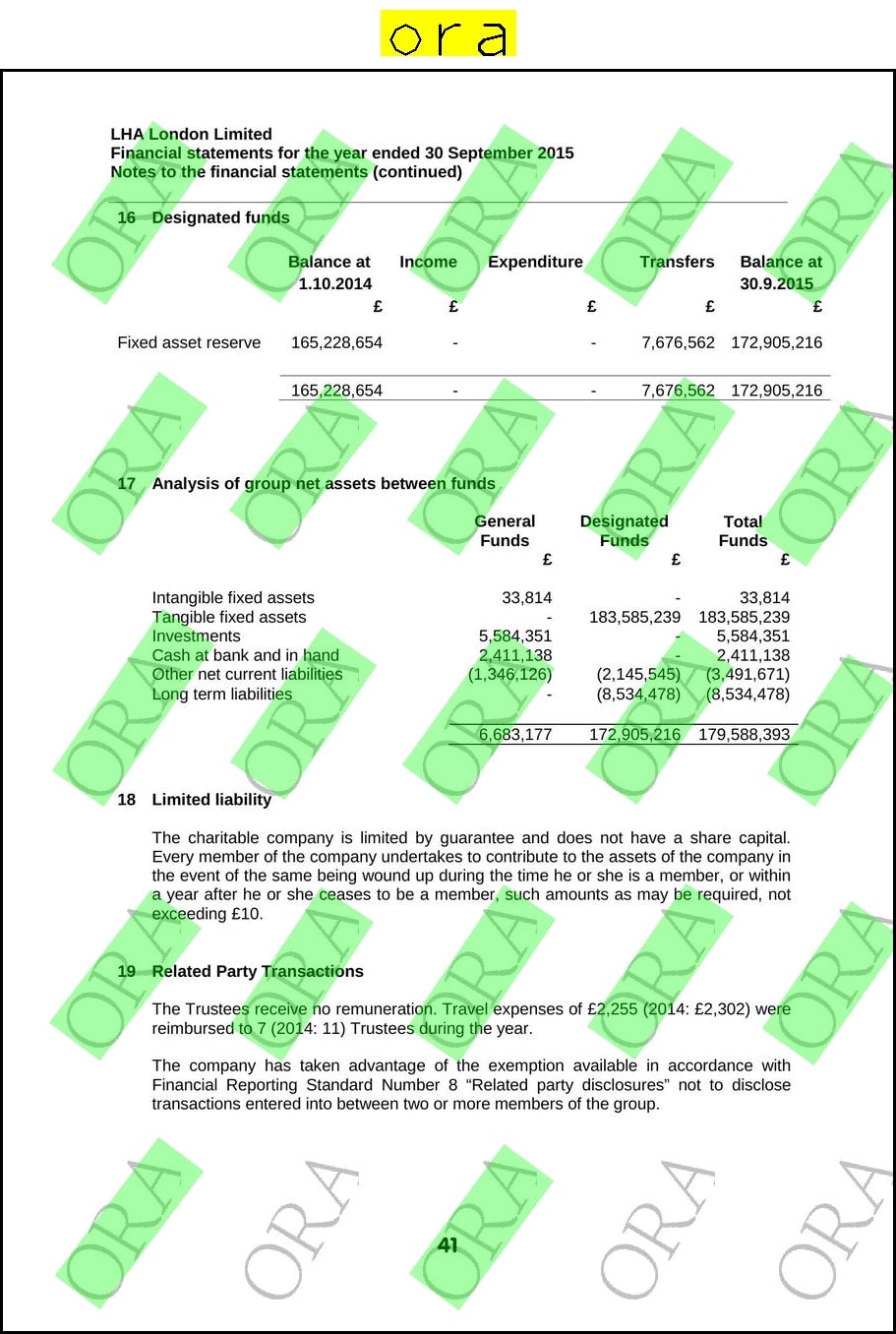}
		\end{tabular}}
	\end{center}
	\caption{\textbf{Qualitative assessment of $\mathcal{W}$extract against comparing baselines.} Notice that our approach provides the highest coverage of watermark text patterns across the document page and generates the closest text prediction w.r.t. the depicted text.}
	\label{fig:visual_result_comparison}
\end{figure*}

\section{Experiments}
We experimented on our own augmented dataset, \textbf{K-Watermark}, created using the $\mathcal{W}$\textbf{render} algorithm. The dataset images encode a high amount of variability in terms of visual elements (containing simple text document pages and visually rich document pages depicting images, logos or table structures), text density and semantics to represent a principled evaluation framework. To the best of our knowledge, we are the first to propose an approach which retrieves exclusively watermark text content. We adapted the text spotting baselines for a fair comparison. We compared against text detection approaches adapted for watermark text detection, combined with text recognition approaches.

\begin{table*}[t]
	\centering
	\scalebox{0.8}{
	\begin{tabular}{@{}c | c | c | c | c | c | c | c }
	    \backslashbox{Detection}{Recognition} & RobustScanner \cite{yue2020robustscanner}  & MASTER \cite{lu2021master} & SATRN \cite{lee2020recognizing} & PARSeq \cite{bautista2022parseq} & ABINet \cite{fang2021read} & UNITS \cite{kil2023towards} & $\mathcal{W}$\textbf{extract} \\
		\hline
	    TextSnake \cite{long2018textsnake} & $0.60 \pm 0.10$ & $0.60 \pm 0.09$ & $0.61 \pm 0.11$ & $0.55 \pm 0.12$ & $0.75 \pm 0.08$  &   \\
		\cline{1-6}
		DBNet++ \cite{liao2022real} & $0.56 \pm 0.13$ & $0.56 \pm 0.11$ & $0.56 \pm 0.12$ & $0.55 \pm 0.14$ & $0.69 \pm 0.14$ & $0.70 \pm 0.09$  & $\mathbf{0.79} \pm \mathbf{0.02}$ \\
		\cline{1-6}
		TCM \cite{yu2023turning} & $0.51 \pm 0.19$ & $0.53 \pm 0.17$ & $0.53 \pm 0.21$ & $0.58 \pm 0.15$ & $0.67 \pm 0.21$ & \\
	\end{tabular}
	}
	\caption{\textbf{Watermark Text Spotting Results on K-Watermark Test Set.} We report the mean character accuracy metric $\pm$ standard deviation at document level by comparing our proposed $\mathcal{W}$\textbf{extract} (\textit{rightmost column}) against different combinations of  state-of-the-art text-spotting methods (\textit{rows}) and text recognition frameworks (\textit{columns}). $\mathcal{W}$\textbf{extract} and UNITS \cite{kil2023towards} are placed on a stand-alone column as they have end-to-end flows w.r.t. the retrieval of watermark text. Our approach outperforms the overall comparing baselines by a large margin. The second best performing one with close performance involves ABINet \cite{fang2021read} recognition. Moreover, our method demonstrates consistent performance at document level as it has the lowest standard deviation \textit{i.e.}, $0.02$. }
	\label{tbl:recognition_results}
\end{table*}

\subsection{Text Detection} \label{sec:detection}
The detection component was trained using an SGD optimiser with a constant learning rate of $5 \times 1e^{-3}$, an epoch of warmup and a batch size of $16$. We initialise the Faster-RCNN with feature pyramid networks \cite{Lin_2017_ICCV} using weights obtained through COCO pretraining \cite{lin2014microsoft}. We make use of $\mathbf{\mathcal{W}render}$ dynamically: at each epoch the same document page has different patterns rendered. To maintain a reasonable training time, we down-scale the images so that the longest edge has between $800$ and $1024$ pixels. Detailed experimental results on the \textbf{K-Watermark} test set are reported in Table \ref{tbl:detection_results}. We evaluate the watermark text detection component by computing standard rotated object detection metrics (\textit{i.e.}, average precision and average recall). 

We report the mean average precision (mAP) and mean average recall (mAR) with an intersection over union (IoU) overlap threshold between $0.5$ and $0.95$. Additionally, we provide evaluation at fixed IoU thresholds of $0.5$ and $0.75$. We compare with the state-of-the-art approaches for text detection \cite{long2018textsnake,baek2020character,liao2022real,yu2023turning}, but also with the detection results of an end-to-end text spotting method \cite{kil2023towards}. For a fair comparison, we have two baseline comparison setups. The first baseline setup (Table \ref{tbl:detection_results} lines $2$-$6$) is to use off-the-shelf weights trained on public scene text detection datasets and recover all detection instances of text from the input image (including \textit{e.g.}, document text, logo text, watermark text). Firstly, we apply those methods both on the normal document images (pre-watermark insertion) and on the watermarked document images (post-watermark insertion), retaining the set of predictions which only occur on the latter. 

Even with this advantageous evaluation setup, our pipeline outperforms these baselines by a large margin ($\approx 50$ points in mAP). In terms of recall, the gap is smaller due to the fact that the considered detection baselines pick up all types of text patterns from the document page, including watermark text patterns.

The second baseline heuristic (Table \ref{tbl:detection_results} lines $7$-$10$) is to fine-tune the baselines on the K-Watermark training set, thus specializing the framework in retrieving watermark text bounding boxes solely. The performance is far superior compared to their off-the-shelf version, however inferior to our $\mathcal{W}$\textbf{extract} approach. We were unable to apply this comparison heuristic on \cite{baek2020character} as the authors do not provide the training code. Prior to exploring the text recognition branch, we analysed the impact of using various backbones as alternatives to the \texttt{ResNet}$50$, however, as it is noticeable in Table \ref{tbl:ablation_backbone}, other transformer-based \cite{liu2021swin,li2022exploring} alternatives performed poorer and are more memory-intensive due to the increased number of parameters.

Moreover, we perform an ablation study by training $\mathcal{W}$\textbf{extract} with and without the $\mathcal{L}_\texttt{VAR}$ (lines $11$ and $12$ from Table \ref{tbl:detection_results}). The $\mathcal{L}_\texttt{VAR}$ usage results in an improvement of $2$ mAP and mAR points, as it leverages the consistency of the watermark patterns in terms of dimension and orientation across the page.
In Figure \ref{fig:visual_result_b}, \textit{row} $1$, we illustrate some sample visual results of $\mathcal{W}$\textbf{extract} on \textbf{K-Watermark} dataset. The method generates consistent watermark predictions across the page, which are in part influenced by the $\mathcal{L}_\texttt{VAR}$ loss component. Also, it demonstrates robustness with respect to diagonally placed document text, high occlusion degree against background text or other visual elements such as logos, pictures or document page symbols. In Figure \ref{fig:visual_result_b}, rows $2$ and $3$, we visually demonstrate the robustness of our approach by testing on pre-generated watermark documents. This validates that our training procedure does not bias towards the $\mathcal{W}$\textbf{render} procedure and it is able to detect and recognize watermark text patterns from other sources. In addition, in Figure \ref{fig:failures_detection} we showcase examples when our framework fails to capture all the watermark text instances or when various visual artefacts are recovered as faulty predictions. 

\subsection{Text Recognition}
The text recognition head was trained jointly with a pre-trained watermark detection backbone using AdamW optimiser \cite{LoshchilovH19} with an inverse square root schedule that had a starting learning rate of $10^{-5}$, $\beta_1 = 0.9$, $\beta_2 = 0.98$, $\epsilon = 10^{-9}$, $\gamma = 0.8$ and batch size of $8$.
Before the joint training, we performed a pre-training step for the $\mathrm{\Psi}_{\textrm{TXT}}$ decoder by freezing the detection backbone and using boxes corresponding to ground truth watermarks. It allowed the decoder to learn the semantic distribution of the watermark texts and helped with the stability of training.
In order to assess the text recognition performance, we use only the fine-tuned versions of detection baselines, since these are the top scoring ones from Table \ref{tbl:detection_results} and apply state-of-the-art text recognition approaches \cite{yue2020robustscanner,lu2021master,lee2020recognizing,fang2021read}. Since $\mathcal{W}$\textbf{extract} produces a single character sequence prediction per document, and the baselines we work with generate one for each watermark instance, we apply a majority voting aggregation procedure to produce a single sequence. Specifically, we retrieve the character sequence which occurs most frequently among the list of predicted character sequences. The numerical results averaged at document level are illustrated in Table \ref{tbl:recognition_results}. To measure the text recognition performance, we calculate the character accuracy, \textit{i.e.}, 

\begin{align*}
	1 - \frac{\# \texttt{CHAR}_\texttt{\textbf{SUBST}} + \# \texttt{CHAR}_\texttt{\textbf{DEL}} + \# \texttt{CHAR}_\texttt{\textbf{INS}}}{\# \texttt{CHAR}_\texttt{\textbf{TOTAL}}}
\end{align*}

\noindent where $\# \texttt{CHAR}_\texttt{\textbf{SUBST}}$, $\# \texttt{CHAR}_\texttt{\textbf{DEL}}$, $\# \texttt{CHAR}_\texttt{\textbf{INS}}$ and $\# \texttt{CHAR}_\texttt{\textbf{TOTAL}}$ represent the number of character substitutions, number of character deletions, number of character insertions and total number of characters, respectively.

The only baseline with a comparable performance is \cite{fang2021read} combined with either \cite{long2018textsnake} or \cite{liao2022real}. Our assumption is that the cause of this is the language model component, which adjusts the prediction to resemble a semantically meaningful word entry.  
One interesting aspect is the fact that although \cite{liao2022real} achieves superior detection results compared to \cite{long2018textsnake}, in terms of recognition it is inferior. Based on our empirical observations, the method of \cite{liao2022real} captures more instances of watermark text patterns. However when combined with the majority voting heuristic, it induces more noise in the recognition pipeline, thus creating a slight performance gap. Although the model of \cite{long2018textsnake} has a poorer detection performance, it captures the most relevant patterns w.r.t. the recognition task, thus less noise and better recognition performance. The method proposed in \cite{kil2023towards} is the only to provide an end-to-end flow which retrieves the watermark text pattern instances across the document page and their semantic meaning. We were able to fully fine-tune it across our proposed dataset. However, we still had to apply the majority voting procedure, as the method outputs a text prediction for each watermark box. 

To better understand the limitations and the advantages of our method, we conduct a detailed analysis of the recognition performance. In Figure \ref{fig:plots_recognition} we show two performance plots highlighting how the character accuracy is impacted by the fadedness (\emph{right}) and the angular orientation (\emph{left}) of the watermark text. We combined the text recognition pipelines with the best text detection approach for easier visualisation purposes. Our $\mathcal{W}$\textbf{extract} is able to achieve the highest performance in situations of very high fadedness (visibility degree of $0.1$) or with almost vertical text orientation (angle of $\approx 90$ degrees with respect to the horizontal axis). Apart from this, we manifest constant high performance across all angular orientations and visibility degrees. This observation is in line with the smallest standard deviation showcased by our approach in Table \ref{tbl:recognition_results}. In Figure \ref{fig:visual_result_comparison} we showcase visual comparisons between our proposed approach and UNITS \cite{kil2023towards} and TCM \cite{yu2023turning}-ABInet\cite{fang2021read} baselines. Our approach demonstrates maximum coverage of the watermark text pattern and robustness w.r.t. the orientation, fadedness, font type and high overlap against document text.

Moreover, in Table \ref{tbl:ablation_study} we perform an ablation study to understand the impact of the local and global feature representations. We re-trained the recognition pipeline with all combinations of global and local feature representations. The global features are primarily specialized for the object detection tasks (classification and bounding box regression), thus using them solely performs poorer with respect to the text recognition task. 

\begin{table}[h]
    \resizebox{\columnwidth}{!}{
	\begin{tabular}{@{}c | c | c | c@{}}
		$\mathcal{W}$extract Feature Type & $[\mathrm{\Theta}_\texttt{LOCAL}; \mathrm{\Theta}_\texttt{GLOBAL}]$ &  $[\mathrm{\Theta}_\texttt{LOCAL}]$ & $[ \mathrm{\Theta}_\texttt{GLOBAL}]$ \\
		\hline
		Character Accuracy & $\textbf{0.791}$ & $0.721$ & $0.479$ \\
	\end{tabular}}
	\caption{\textbf{Ablation on importance of \emph{global} vs \emph{local} features.} This study proves the importance of using the combined \emph{global} and \emph{local} information when determining the watermark text - they complement each other with respect to the task of text recognition. The usage of \emph{global} features alone underperforms severely, as they are not able to overcome the local noise induced by overlapping with document text and high fadedness factors.}
	\label{tbl:ablation_study}
\end{table}

\section{Conclusions}
We present a principled benchmark for watermark text spotting from documents together with an end-to-end solution for detecting watermark text patterns and recognizing the depicted text. Our key novelty is a hybrid joint watermark text pattern detection and recognition system leveraging a global and local hierarchical attention mechanism robust to occlusion, low visibility and various degrees of document text densities which is constrained by a variation minimisation loss term. This is a niche and important problem with high potential impact, which can be considered as an extension for any OCR. No previous existing methodological approaches exist, thus we compared our proposed watermark text spotting pipeline, $\mathcal{W}$\textbf{extract}, against strong text-spotting baselines specifically adapted for our problem, outperforming them by a large margin.
\pagebreak
\bibliography{main}


\end{document}